\pdfoutput=1
\documentclass[11pt]{article}

\usepackage[final]{acl}

\usepackage{times}
\usepackage{latexsym}

\usepackage[T1]{fontenc}
\usepackage[utf8]{inputenc}

\usepackage{microtype}

\usepackage{inconsolata}

\usepackage{graphicx}

\usepackage{enumitem}

\usepackage{rotating}  % Needed for the sidewaystable environment
\usepackage{amssymb}% http://ctan.org/pkg/amssymb
\usepackage{pifont}% http://ctan.org/pkg/pifont
\newcommand{\cmark}{\ding{51}}%
\newcommand{\xmark}{\ding{55}}%
\usepackage{color,soul}
\usepackage{lscape}
\usepackage{textcomp}
\usepackage{multirow}
\usepackage{graphicx}
\usepackage{xcolor}
\usepackage{array}
\usepackage{booktabs}
\usepackage{soul}
\newcommand{\ctext}[3][RGB]{%
  \begingroup
  \definecolor{hlcolor}{#1}{#2}\sethlcolor{hlcolor}%
  \hl{#3}%
  \endgroup
}

\usepackage{rotating}
\usepackage{caption} 
\usepackage{colortbl}
\usepackage{multirow}
\usepackage{tabularray}

\definecolor{green}{RGB}{153,255,153}
\definecolor{hred}{RGB}{255,153,153}
\definecolor{hyellow}{RGB}{255,255,153}
\definecolor{orange}{RGB}{255,165,0}
\definecolor{lightblue}{RGB}{173,216,230}

\newcommand{\highlightred}[1]{\sethlcolor{red!40}\hl{#1}}
\newcommand{\highlightblue}[1]{\sethlcolor{blue!40}\hl{#1}}
\newcommand{\highlightorange}[1]{\sethlcolor{orange!40}\hl{#1}}
\newcommand{\highlightgreen}[1]{\sethlcolor{green!50}\hl{#1}}
\newcommand{\highlightviolet}[1]{\sethlcolor{violet!40}\hl{#1}}

\usepackage{enumitem}
\setitemize{noitemsep,topsep=0pt,parsep=0pt,partopsep=0pt}

\usepackage{float}
\usepackage{longtable}
\usepackage{tabularx, booktabs}

\usepackage{subcaption}

\title{A Modular Taxonomy for Hate Speech Definitions and Its Impact on Zero-Shot LLM Classification Performance}
\author{
 \textbf{Matteo Melis\textsuperscript{1}},
 \textbf{Gabriella Lapesa\textsuperscript{2,3}},
 \textbf{Dennis Assenmacher\textsuperscript{2}}
\\
 \textsuperscript{1}Department of Linguistics, Cognitive Science and Semiotics - Aarhus University, \\ \textsuperscript{2}GESIS - Leibniz Institute for the Social Sciences  \\
\textsuperscript{3}Heinrich-Heine University Düsseldorf \\
\textsuperscript{1}\texttt{mmls@cc.au.dk}, \textsuperscript{2}\texttt{first.last@gesis.org} 
}
\begin{document}
\maketitle
\begin{abstract}
%Detecting harmful content is a crucial task in the landscape of NLP applications for Social Good, with hate speech being one of its most dangerous forms. But what do we mean by hate speech and how can we define it? Given the common use of LLMs in zero-shot settings for these tasks: how does prompting different definitions of hate speech affect their performance? This paper addresses these two research challenges and, therefore, its contribution is twofold. At the theoretical level, to clarify the ambiguity related to the notion of hate speech, we collect and analyze existing definitions of hate speech in the literature, and organize them in a taxonomy composed of 14 conceptual elements (i.e., building blocks of the definitions), which provide different pieces of information related to the construct definition of hate speech (e.g., mention of the target of hate). At the experimental level, we exploit the conceptual fine-graininess of the proposed taxonomy to carry out a systematic zero-shot evaluation of 3 LLMs, in terms of performance and consistency over 3 datasets which represent three different data types (synthetic, human-in-the-loop, and real-world).  We find that choosing different definitions impacts model performance, but this effect is not consistent across all architectures.

Detecting harmful content is a crucial task in the landscape of NLP applications for Social Good, with hate speech being one of its most dangerous forms. But what do we mean by hate speech, how can we define it, and how does prompting different definitions of hate speech affect model performance? The contribution of this work is twofold. At the theoretical level, we address the ambiguity surrounding hate speech by collecting and analyzing existing definitions from the literature. We organize these definitions into a taxonomy of 14 Conceptual Elements—building blocks that capture different aspects of hate speech definitions, such as references to the target of hate (individual or groups) or of the potential consequences of it. %Next, we use the building blocks in the taxonomy to build a collection of hate speech definitions that is organized in terms of increasing informativeness and specificity (i.e., number and type of encoded conceptual elements).  
At the experimental level, we employ the collection of definitions in a systematic zero-shot evaluation of three LLMs, on three hate speech datasets representing different types of data (synthetic, human-in-the-loop, and real-world). We find that choosing different definitions, i.e., definitions with a different degree of specificity in terms of encoded elements,  impacts model performance, but this effect is not consistent across all architectures.

%Detecting harmful content is a crucial task in the landscape of NLP applications for Social Good. But what do we mean by Hate Speech? and how does prompting with different definitions of Hate Speech affect the performance of LLMs in the zero-shot setting? This paper addresses these two research challenges and, therefore, its contribution is twofold. At the theoretical level, to clarify the ambiguity related to the construct of Hate Speech, we collect and analyze existing definitions of Hate Speech in the literature, and organize them in a taxonomy composed of 14 conceptual elements (i.e., building blocks of the definitions), which provide different pieces of information related to the construct definition of Hate Speech (e.g., mention of the target of hate). At the experimental level, we exploit the conceptual fine-grainedness of the proposed taxonomy to carry out a systematic evaluation of zero-shot models, in terms of performance and consistency over three datasets which represents three different data-type (synthetic, human-in-the-loop, real world).  We find that choosing different definitions impacts model performance, but this effect is not consistent across all architectures.

\end{abstract}

\section{Introduction}\label{sec:Intro}

In a world that is becoming increasingly online-based, detecting harmful content, specifically Hate Speech (HS), is crucial for maintaining the integrity of the democratic discourse and freedom of speech \citep{Kiritchenko, Tsesis}.
The advent of Large Language Models (LLMs) paved the way for a variety of new methods for detecting \cite{Roy} and countering HS \cite{bonaldi-etal-2023-weigh}, and for  the creation of new artificial benchmarking data \cite{jin-etal-2024-gpt-hatecheck, sen-etal-2023-people}.

In particular, novel methods for classifying harmful content diverge from conventional supervised learning that relies on input/output pairs, but uses only predetermined prompts without examples \cite{plaza-del-arco-etal-2023-respectful} or adding further information on the task \cite{Roy}.

A crucial role in refining prompts for zero-shot classification is played by the definition of the target construct, i.e., in the focus of this paper, \textit{the definition of hate speech}.\footnote{A  construct is defined as “an idea or theory containing various conceptual elements, typically one considered to be subjective and not based on empirical evidence"\href{https://languages.oup.com/google-dictionary-en/}{(Oxford Languages)}} As typical of social constructs of the social sciences, the definition of HS is ambiguous \cite{plaza-del-arco-etal-2023-respectful,Waseem} and cannot be easily framed in a static dimension. This is a relevant issue for the community, because it affects the interoperability of resources annotated at high cost, and the comparability of the results (and insights) drawn from their modeling, when, for example, different definitions are used for equivalent concepts \cite{Fortuna}. 

The contribution of our work is twofold: conceptual/theoretical and experimental.

At the conceptual level, we contribute to structuring the conceptual landscape of HS by collecting and qualitatively organizing various definitions for hate speech. The goal of this analysis is to identify a set of Conceptual Elements (CEs), i.e., the conceptual building blocks present in the definitions, which encode their key dimensions. For instance, all definitions highlight its problematic nature (CE = Problematic Content) and specify that the target is an individual or group (CE = Target). However, only some definitions include potential consequences of hate speech (CE = Possible Implications) or acknowledge that it can be implicit (CE = Implicit Hate). This taxonomy serves as a scaffold for constructing and analyzing definitions, which we believe is a novel and practically valuable contribution to both the NLP and social science communities.

\begin{figure}[h]
    \centering
    \includegraphics[width=\columnwidth]{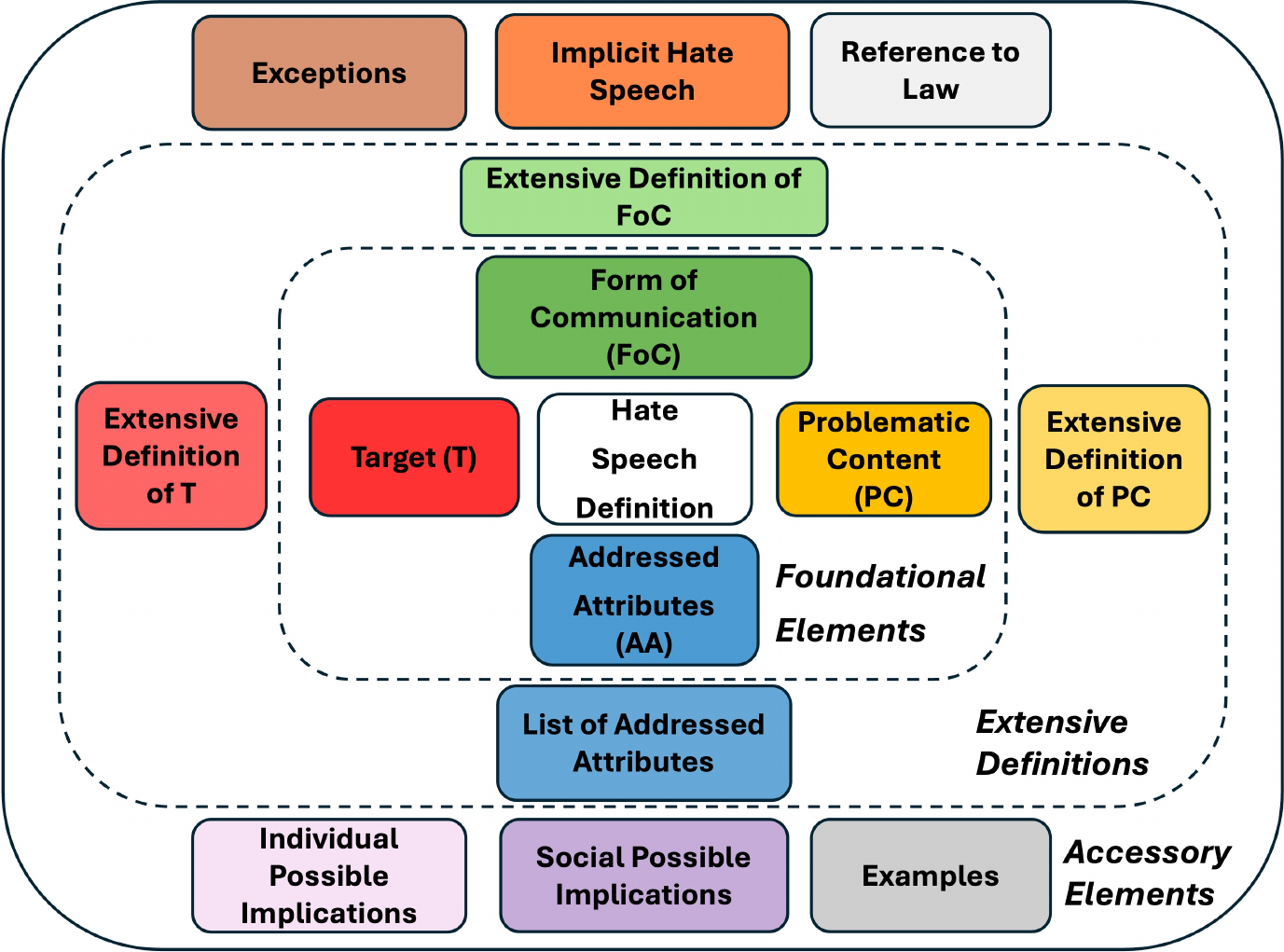}
    \caption{Taxonomy for Hate Speech definitions. To ease the readability of the work, the  Conceptual Elements are color-coded. Refer to Appendix \ref{app:taxi} for the full size figure.}
    \label{fig:taxi_main}
\end{figure}

With these Conceptual Elements, we create a three-layer taxonomy (Fig. \ref{fig:taxi_main}) that we complement with a curated collection of definitions that arise from their combination. The collection of definitions can be seen as a structured, modular summary of the original set of definitions we reviewed and constitute a resource that we make available to the community for further experimentation.

The second contribution is experimental: starting from the idea that LLMs already encode extensive knowledge due to their pre-training and instruction-tuning \cite{Zhang} we employ our definitions from the collection to carry out zero-shot prompting experiments on three hate-speech datasets, representing different data types: Hate\-Check (\citealp{rottger2021hatecheck}, synthetic), Learning from the Worst (\citealp{vidgen-etal-2021-learning}, human-in-the-loop) and Measuring Hate Speech (\citealp{sachdeva2022measuring}, real-world examples). We employ  three different LLMs: \texttt{LLama-3}, \texttt{Mistral} and \texttt{Flan-T5}. We conduct an in-depth error analysis and exploit the Hate\-Check fine-grained annotation regarding types of hate.\footnote{The code can be found at: \href{https://github.com/matteo-mls/Modular-Taxonomy-for-Hate-Speech-Definitions}{https://github.com/matteo-mls/Modular-Taxonomy-for-Hate-Speech-Definitions}}

Our results demonstrate the usefulness of our modular approach to build Hate Speech definitions as prompts for zero-shot classification. 
We find that varying construct definitions affects model performance, but this effect is not consistent across all model architectures and datasets. In some cases, more detailed definitions reduce false negatives, in others they primarily decrease false positives; more specifically, our error analysis shows that more detailed definitions improve performance in cases requiring nuanced distinctions between hate categories (i.e., Implicit Hate).

\section{Related Work}

\paragraph{Zero-shot prompting: general evaluation issues and application to HS}

With no need for computationally expensive fine-tuning, zero-shot prompting allows researchers to  "just ask" a LLM to perform a task (e.g., classification). Unsurprisingly, this strategy is very frequently employed in scenarios with low computational power (e.g., social scientists with no access to fine-tuning infrastructures). The evaluation challenges related to zero-shot prompting have recently been explored in depth by \citet{Beck}, who reported differences in robustness and sensitivity when prompting diverse socio-demographic information along with the evaluated tasks. 

Similar strategies have been observed in HS detection. Prompting LLMs with information on the task different from examples has improved performance in detecting HS \cite{Roy, plaza-del-arco-etal-2023-respectful}. Promising strategies involve prompting rationales or Chain-of-Thoughts alongside the task in zero-shot learning (ZSL), few-shot learning, or fine-tuning \cite{Yang,Nghiem}. These approaches have shown that in-context learning, particularly in the context of ZSL, is a worthwhile direction to explore \cite{Ziems}.

\paragraph{Construct Definition for HS detection}

Previous work explored how construct definitions can be utilized to obtain dataset-specific model-generated rationales \cite{Nghiem}. Other researchers explored how using a definition for an annotation task leads to more consistent answers among both human annotators \cite{Ross} and LLMs \cite{Li}, affecting also their performance. \citet{Roy} investigated the effects of prompting different information (e.g., target, explanation) also among the HS construct definition. Their findings suggest definition-prompting led to mixed results, sometimes worsening and sometimes improving performance across various datasets.

Choosing an adequate definition to describe the construct of HS is challenging. There are overlapping and duplicate definitions \cite{Fortuna}, and sometimes conceptually different constructs are conflated, such as
HS and Offensive Language (OL) \cite{Davidson}. In other cases, different constructs are put under the same umbrella, for example: HS, abusive and discriminatory language \cite{Goldzycher}.  %This confusion in the field indicates little agreement on a shared and easily communicable definition of HS. 
Furthermore, there seems to be minimal effort towards providing a more standardized definition. To the best of our knowledge, only \citet{Khurana} propose 5 criteria, taking also into account a legal perspective. In this work, we propose a taxonomy composed of 14 Conceptual Elements of which only three overlap with \citet{Khurana}.\footnote{Two of them are what we call the "Target” and the “Problematic Content “ and the third is “Possible Implications”.}

%\paragraph{Research gap} In this work we will focus on two research gaps, first we aim to clarify the landscape of HS definition with a taxonomy of its composing elements. Second, we investigate in depth the effect of definition prompting for HS classification.

\section{A Taxonomy for Hate Speech Definitions}

%The taxonomy we propose takes its Conceptual Elements (CEs) as coordinates. These 14 CEs are organized into three layers,which reflects the level of information that is added by each CE, respectively: base/foundational information (Foundational Elements), further information on the construct of HS (Extensive Definitions), and different information from the construct of HS (Accessory Elements). 

\subsection{Procedure}

We reviewed the HS literature over a substantial time span (2000–2021), with a focus on works that operationalized a definition to create datasets or corpora (9 definitions). Additionally, we selected two definitions from conceptual studies on HS \cite{tsesis2002destructive,nockleby2000hate} and two from works on HS detection \cite{mandl2021overview,gao2017recognizing}.

In total, we collected 20 HS definitions (see Appendix \ref{app:definitions}) from the following sources:

\begin{itemize}[leftmargin=*]
    \item 13 definitions from literature \cite{sachdeva2022measuring,vidgen-etal-2021-learning,mandl2021overview,rottger2021hatecheck,basile2019semeval,gibert2018hate,founta2018large,Davidson,gao2017recognizing,nobata2016abusive,warner2012detecting,tsesis2002destructive,nockleby2000hate};
    \item 3 definitions from social networks policies (Twitter/$\mathbb{X}$, Facebook, Youtube);
    \item 2 definitions automatically generated by LLMs (ChatGPT, Gemini);
    \item 2 definitions from official documents (\href{https://www.un.org/en/hate-speech/understanding-hate-speech/what-is-hate-speech}{UN Strategy and Plan of Action on Hate Speech}, Code of Conduct between European Union Commission and companies, \citealp{wigand2017speech});
\end{itemize}

Using these definitions, we inductively identified 14 CEs (building blocks of the HS construct) which we organize in three layers (see Appendix \ref{app:taxi} for a visual representation).

Defining the taxonomy presents two key challenges. First, distinguishing Offensive Language (OL) from Hate Speech (HS) is complicated by a confounding effect noted by \citet{Davidson} and \citet{Waseem}, where OL and HS overlap. We clarify that while OL can exist without being HS, any content classified as HS must also be considered OL. Second, avoiding circular definitions is crucial (i.e., a definition that relies on another definition to be understood). While ‘protected groups’ are often used to differentiate HS from OL, and this approach has legal relevance \cite{Khurana}, using this as a defining criterion, would mean defining HS by using another definition, which varies in relation to culture, laws and people's sensitivity. Definitions in prior work often rely on the identification of protected groups \cite{gibert2018hate}. However, based on the researcher's choice, the protected groups can be listed in the definition,\footnote{In what we later define as List of Addressed Attributes (LAA)} but we do not recommend to use them as defining factor. Instead, our approach shifts the focus from only enumerating categories to explicitly describing the dynamic of attacking a target based on some "inherent characteristics that are attributed to that group and shared among its members".

%we propose explicitly stating that HS conveys malevolent intent toward a group or an individual based on their inherent (or perceived) characteristics.

\subsection{Taxonomy}

Below, we describe each of the three layers of CEs. A detailed description of each CE can be found in Appendix \ref{app:CE_def}. Table \ref{tab:Ce_tab} illustrates the different Conceptual Elements and corresponding abbreviations.

\paragraph{Foundational Elements}

We label the most common and therefore most important CEs as Foundational Elements, being essential for constructing a meaningful definition of HS. These elements include: Form of Communication (\ctext[RGB]{176, 235, 180}{FoC}), Target (\ctext[RGB]{255, 120, 120}{T}), and Problematic Content (\ctext[RGB]{246, 241, 147}{PC}).

However, considering the challenge of distinguishing HS from OL \cite{Davidson, Waseem}, we added another Foundational Element, the Addressed Attributes (\ctext[RGB]{147, 198, 231}{AA}). This element reflects the explicit relationship between the target and the inherent or perceived characteristics being attacked (e.g., attacking someone based on the belief they follow a specific religion).

These four Conceptual Elements—FoC, T, PC, and AA—together form the basis of a foundational HS definition, which from now on we will refer to as the Hate Speech Base (HSB) definition, and represent the minimal conceptual units that are consistently present in almost all hate speech definitions

%The remaining 10 Conceptual Elements are divided into two groups based on the type of information they contribute to the definition of HS.

\paragraph{Extensive Definitions of the Foundational Elements} Within the second layer, four Conceptual Elements provide additional detail about the core components, including: Extensive Definitions of Form of Communication (\ctext[RGB]{224, 251, 226}{EDFoC}), Target (\ctext[RGB]{255,180,180}{EDT}), and Problematic Content (\ctext[RGB]{255,255,190}{EDPC}), as well as the List of Addressed Attributes (\ctext[RGB]{204,229,255}{LAA}). These CE capture richer or more granular information about the same dimensions present in the previous layer.

\paragraph{Accessory Elements} The remaining six elements are categorized in the third layer and provide different information from the core components of the construct, in other words, new information: social Possible Implications (\ctext{255,170,255}{sPI}), individual Possible Implications (\ctext{255,170,255}{iPI}), Exceptions (\ctext{208,147,112}{Exc}), Implicit Hate Speech (\ctext{255,153,51}{IHS}), Examples (Exa), Reference to Laws (Law).
\\
%The remaining 10 Conceptual Elements are divided into two groups based on the type of information they contribute to the definition of HS. The first group comprises \textit{Extensive Definitions} of the Foundational Elements, these provide additional detail about the core components, including: Extensive Definitions of Form of Communication (EDFoC), Target (EDT), and Problematic Content (EDPC), as well as the List of Addressed Attributes (LAA). The second group, the \textit{Accessory Elements}, contains those elements that provide different information from the core components of the construct: social Possible Implications (sPI), individual Possible Implications (iPI), Exceptions (Exc), Implicit Hate Speech (IHS), Examples (Exa), Reference to Laws (Law). 

\begin{table*}[h]
\footnotesize
\centering
\begin{tabular}{p{5cm}p{2cm}p{8cm}}
    \toprule
    \textbf{Conceptual Element} & \textbf{CE} & \textbf{Example in definition} \\
    \hline
    \multicolumn{3}{c}{\textbf{Foundational Conceptual Elements}} \\
    \hline
    \ctext[RGB]{176, 235, 180}{Form of Communication} + \ctext[RGB]{255, 120, 120}{Target} + \ctext[RGB]{246, 241, 147}{Problematic Content} = \textbf{Offensive Language} & \ctext[RGB]{176, 235, 180}{FoC} + \ctext[RGB]{255, 120, 120}{T} + \ctext[RGB]{246, 241, 147}{PC} = \textbf{OL} & Hate Speech is considered \ctext[RGB]{176, 235, 180}{any kind of content} that conveys \ctext[RGB]{246, 241, 147}{malevolent intentions} toward a \ctext[RGB]{255, 120, 120}{group or an individual}. \\
    \hline
    \ctext[RGB]{176, 235, 180}{Form of Communication} + \ctext[RGB]{255, 120, 120}{Target} + \ctext[RGB]{246, 241, 147}{Problematic Content} + \ctext[RGB]{147, 198, 231}{Addressed Attributes} = \textbf{Hate Speech Base} & \ctext[RGB]{176, 235, 180}{FoC} + \ctext[RGB]{255, 120, 120}{T} + \ctext[RGB]{246, 241, 147}{PC} + \ctext[RGB]{147, 198, 231}{AA} = \textbf{HSB} & Hate Speech is considered \ctext[RGB]{176, 235, 180}{any kind of content} that conveys \ctext[RGB]{246, 241, 147}{malevolent intentions} toward a \ctext[RGB]{255, 120, 120}{group or an individual}, \ctext[RGB]{147, 198, 231}{and motivated by inherent characteristics that are attributed to that group and shared among its members}. \\
    \hline
    \multicolumn{3}{c}{\textbf{Extensive Definition of the Foundational Elements (Step 1)}} \\
    \hline    
    \textbf{Hate Speech Base} + \ctext[RGB]{224, 251, 226}{Extensive Definition Form of Communication} & \textbf{HSB} + \ctext[RGB]{224, 251, 226}{EDFoC} & Hate speech is considered any kind of content \ctext[RGB]{224, 251, 226}{or communication expressed using language (written or spoken) or actions,}  that convey malevolent intentions toward a group or an individual, and motivated by inherent characteristics that are attributed to that group and shared among its members. \\
    \hline
    \textbf{Hate Speech Base} + \ctext[RGB]{255,180,180}{Extensive Definition Target} & \textbf{HSB} + \ctext[RGB]{255,180,180}{EDT} & Hate speech is considered any kind of content that conveys malevolent intentions toward a group or an individual \ctext[RGB]{255,180,180}{which is, or thought to be, a member of that group}, and motivated by inherent characteristics that are attributed to that group and shared among its members. \\
    \hline
    \textbf{Hate Speech Base} + \ctext[RGB]{255,255,190}{Extensive Definition Problematic Content} & \textbf{HSB} + \ctext[RGB]{255,255,190}{EDPC} & Hate speech is considered any kind of content that conveys malevolent intentions \ctext[RGB]{255,255,190}{such as statements of inferiority, aversion, cursing, calls for exclusion, threaten, harass or violence}, toward a group or an individual, and motivated by inherent characteristics that are attributed to that group and shared among its members. \\
    \hline
    \multicolumn{3}{c}{\textbf{Accessory Elements (Step 2)}} \\
    \hline
    \textbf{Hate Speech Base} + \ctext[RGB]{204,229,255}{List of Addressed Attributes} & \textbf{HSB} + \ctext[RGB]{204,229,255}{LAA} & Hate speech is considered any kind of content that conveys malevolent intentions toward a group or an individual, and motivated by inherent characteristics that are attributed to that group and shared among its members \ctext[RGB]{204,229,255}{such as race, color, ethnicity, gender, sexual orientation, nationality, religion, disability, social status, health conditions, or other characteristics}.\\
    \hline
    \textbf{Hate Speech Base} + \ctext{255,170,255}{Possible Implications} & \textbf{HSB} + \ctext{255,170,255}{PI} & Hate speech is... \ctext{255,170,255}{The outcome of Hate Speech could be the promotion of division among people, undermining of social cohesion in communities, inciting others to commit violence or discrimination, and could have consequences for individuals’ health and safety}. \\
    \hline
    \textbf{Hate Speech Base} + \ctext{208,147,112}{Exception} & \textbf{HSB} + \ctext{208,147,112}{Exc} & Hate speech is... \ctext{208,147,112}{However, even if it is offensive, it is not considered Hate Speech any content that attacks a person’s personality traits, ideas, or opinions}. \\
    \hline
    \textbf{Hate Speech Base} + \ctext{255,153,51}{Implicit Hate Speech} & \textbf{HSB} + \ctext{255,153,51}{IHS} & Hate speech is... \ctext{255,153,51}{Hate Speech can also be implicit, portrayed as an indirect or coded language that uses Irony, Stereotypes, or Misinformation}. \\
    \bottomrule
  
\end{tabular}
\captionsetup{font=footnotesize}    
\caption{Colour coded Conceptual Elements and examples in the definitions prompted in the HS detection task.}
\label{tab:Ce_tab}
\end{table*}

\subsection{Building definitions from the taxonomy}

Based on the CEs and their modular arrangement within the taxonomy, we generated a collection of definitions by recombining them according to the criteria outlined below.

First, we created content reflecting each CE. For example, the CE: \textit{Target} would be mapped in the natural language expression "toward a group or an individual" while the corresponding, more informative CE: \textit{Extensive Definition of Target} would map into "toward a group or an individual" followed by "which is thought to be a member of that group". 

Second, we combined these elements to create definitions with varying conceptual compositions, aiming to represent different levels of informativeness (level of details of the definition\footnote{We assume that adding more Conceptual Elements leads to higher level of detail}) and types of information (i.e., the specific mention of implicit HS). When combining the CEs to create definitions, we made sure they would not differ in style or wording: for instance, the textual span representing the CE Target is exactly the same in all the definitions.

Table \ref{tab:Ce_tab} lists all the CEs, their abbreviations, and how they are reflected in the definitions we created. In Appendix \ref{app:CE} we showcase the presence or absence of CEs in all the definitions. 
The full set of definitions contained in our collection is reported in Appendix \ref{app:definition_prompted}.

While building the collection of definitions, which was designed for the goal of prompting, we consolidated various forms of potential implications into a single CE: PI (Possible Implications). Additionally, we excluded two CEs—Examples and Reference to Law. The former was omitted to preserve the zero-shot learning (ZSL) condition, as including examples would shift the setup toward few-shot learning. The latter was excluded because assessing models' legal domain knowledge falls outside the scope of this study.

We emphasize that there is no direct one-to-one correspondence between the original definitions used to develop the taxonomy and the definition collection derived from it. Instead, our collection serves as a structured summary of existing definitions, with carefully curated wording to ensure that variation stems solely from different combinations of CEs. This makes it an ideal starting point for the prompting experiments presented in the next section.

\section{Zero-shot prompting}

\subsection{Experimental setup}\label{subsec:dataset}

\paragraph{Datasets} In our zero-shot experiments, we use three different datasets reflecting different data types:

\begin{enumerate}[topsep=0pt,itemsep=0pt,parsep=0pt]
    \item HateCheck \cite{rottger2021hatecheck}: synthetically generated functional test-suite for HS;
    \item Learning from the Worst (LFTW, \citealp{vidgen-etal-2021-learning}): curated collection of challenging HS through a human-in-the-loop process;
    \item Measuring Hate Speech (MHS, \citealp{sachdeva2022measuring}): real-world instances of HS collected from various social media;
\end{enumerate}

These three datasets not only represent different types of data points but also adopt operational definitions of hate speech that align with the Foundational Elements outlined in our taxonomy, ensuring a meaningful and consistent interpretation of hate speech. For reasons of better comparability and to avoid unnecessary computational costs, we randomly sampled from LFTW and MHS the same amount of data-points (3901) with the same distribution among classes (68.16\% Hate Speech, and 31,84\% Not-Hate Speech) of HateCheck. Which we have taken as a reference point due to its structure, which differentiates between all these different functionalities (challenging types of hate), enabling us to investigate them in our error analysis.

%We chose three different datasets for the experimental part: Hate\-Check \cite{rottger2021hatecheck}, Learning from the Worst \cite{vidgen-etal-2021-learning} and Measuring Hate Speech \cite{sachdeva2022measuring}. There are two reasons which motivate the choice of these datasets: a) their operational definitions of HS reflect the Foundational Elements that we suggest in the taxonomy as mandatory to have a meaningful definition of hate speech, b) these datasets reflect three different type of data-points: synthetic (HateCheck), synthetic with humans-in-the-loop (Learning from the Worst) and human-written, collected from social media (Measuring Hate Speech).

%Among these three datasets, given its comprehensive labeling structure that we use in Sec. \ref{subsec:type_of_hate} to investigate error patterns and its nature as a benchmark for Hate Speech, make HateCheck our point of reference. For this reason, in order to avoid useless computational costs and for better comparability, we randomly sample from Learning from the Worst and Measuring Hate Speech datasets the same amount of data-points (3901) with the same distribution among classes (68.16\% Hate Speech, while the remaining 31,84\% are Not-Hate Speech) of HateCheck.

\paragraph{Models}
In our experiments, we employ three open-source, instruction-based LLMs of small to medium sizes from different model families: \texttt{Meta-Llama-3-8B-Instruct}, \texttt{Mistral\-7B-Instruct-v0.2}, and \texttt{Flan-T5-XL}, all sourced from HuggingFace. While for \texttt{LLama-3} and \texttt{Mistral} \cite{Jiang2023Mistral7}, there is no clear information about their pre-training and fine-tuning data, we are only certain that \texttt{Flan-T5} \cite{https://doi.org/10.48550/arxiv.2210.11416} is the only model which was not exposed to any of these particular datasets (though being instruction-tuned on some other hate speech/toxicity datasets, \citealp{wang2022super})\footnote{Only for \texttt{Mistral}, due to numerous instances in which it refused to answer, we used \texttt{outlines} by \citet{willard2023efficient}, a library that allows the user to retrieve a structured generation from the LLMs. This has set the model's temperature to its default value, 0.7, while for the models the temperature was set to 0.95.}

\paragraph{Prompting Strategy}\label{subsec:prompting_strat}

We framed the task as a binary classification task (HS/No Hate Speech (NHS)), keeping the instruction as brief and concise as possible \cite{weber2023mind,chang2024efficient}, Appendix \ref{app:prompts} showcase the resulting prompts. %We also  keep the concept of HS balanced in the task information, repeating it both next to positive and negative instances, resulting in the prompts showcased in Appendix \ref{app:prompts}.

To systematically reduce the number of CE combinations, we followed a two-step approach. First, in \textbf{Step 1}, we refine the definition of Hate Speech Base (HSB), recognizing its central role in our study. We focus on identifying which of the Extensive Definitions of—Form of Communication (FoC), Target (T), Problematic Content (PC), and Addressed Attributes (AA)—
provide the most informative input for the models. 
Secondly, in \textbf{Step 2}, we test the best-performing definition from the Step 1 (highest macro-F1 score) by incorporating additional Accessory Elements: the List of Addressed Attributes (LAA), Possible Implications (PI), Implicit Hate Speech (IHS), and Exceptions (Exc).

For evaluation, we also include: a) each dataset’s respective construct definition (referred to as “Own”), as we expected these definitions to be most reflective of the dataset’s specific data points, and b) a condition in which no definition is given ("NO"), but the model is only asked to classify if the data-point is Hate Speech or not.

\section{Results}
As outlined in Sec. \ref{subsec:prompting_strat}, our experiments followed a two-step approach:\footnote{To ensure stability, each experiment was repeated three times.}

\noindent
\textbf{Step 1}: Which Extensive Definitions provide the most informative refinement of the Hate Speech Base definition?

\noindent
\textbf{Step 2}: How does incorporating additional Conceptual Elements impact the results from Step 1?

\subsection{Step 1: What is the Best Base Refined Definition for Hate Speech?}

\begin{table*}[h]
\scriptsize
\resizebox{\textwidth}{!}{
\begin{tabular}{@{}l|ccc|ccc|ccc!{}}
\toprule
\multicolumn{1}{c|}{\multirow{2}{*}{\textbf{Definitions}}} & \multicolumn{3}{c|}{\textbf{HateCheck}} & \multicolumn{3}{c|}{\textbf{LFTW}} & \multicolumn{3}{c}{\textbf{MHS}} \\
\cmidrule(lr){2-4} \cmidrule(lr){5-7} \cmidrule(lr){8-10}
 & LLama3 & Mistral & FlanT5 & LLama3 & Mistral & FlanT5 & LLama3 & Mistral & FlanT5 \\
\midrule
NO & \textbf{84.82} & \textbf{78.57} & 72.18 & 72.07 & 56.05 & 60.99 & \textbf{75.94} & \textbf{79.12} & 74.21 \\
Own & 76.72 & 75.10 & 75.95 & \textbf{73.86} & 53.83 & 62.43 & 74.17 & 77.10 & 74.79 \\
OL & 77.62 & \textbf{78.57} & 74.40 & 71.75 & \textbf{57.28} & 62.54 & 69.08 & 76.80 & 74.63 \\
HSB & 80.02 & \textit{\underline{78.20}} & 74.91 & 72.63 & 55.78 & 63.66 & 70.72 & 75.81 & 74.30 \\
HSB\_EDFoC & 80.04 & 77.77 & 75.18 & 72.87 & \textit{\underline{55.82}} & 63.41 & 72.00 & \textit{\underline{77.14}} & 75.21 \\
HSB\_EDPC & 78.90 & 76.40 & 75.11 & 71.95 & 54.72 & 63.32 & 73.24 & 76.09 & 74.77 \\
HSB\_EDT & \textit{\underline{80.14}} & 77.17 & \textbf{\textit{\underline{76.29}}} & \underline{73.42} & 54.19 & 63.83 & 72.04 & 75.59 & \textbf{\textit{\underline{75.54}}} \\
HSB\_EDFoC\_EDT & 79.99 & 76.66 & 75.66 & 73.31 & 54.78 & 63.65 & 72.61 & 75.98 & 75.38 \\
HSB\_EDFoC\_EDPC & 80.01 & 76.44 & 74.71 & 72.33 & 55.04 & 62.85 & \textit{\underline{73.99}} & 76.40 & 74.85 \\
HSB\_EDT\_EDPC & 79.59 & 75.58 & 75.97 & 72.52 & 53.17 & 63.76 & 73.77 & 75.76 & 74.58 \\
HSB\_EDFoC\_EDPC\_EDT & 80.06 & 75.54 & 76.21 & 72.64 & 53.46 & \textbf{\textit{\underline{64.19}}} & 73.94 & 76.70 & 75.15 \\
\midrule
\midrule
Pearson Corr. (tokens) & -0.05 & -0.96 & 0.62 & -0.10 & -0.59 & 0.67 & 0.70 & -0.26 & 0.35 \\
\midrule
\midrule
Best Conceptual Elements  & EDT & - & EDT & EDT & EDFoC & EDs & EDFoC\_EDPC & EDFoC & EDT \\
\bottomrule
\end{tabular}}
\caption{Step 1, F1-macro: In \textbf{bold} the highest score, the \underline{underlined} score is the chosen one for the second step. The Correlation Coefficients do not consider the condition without definition (NO). (Own = Definition of the dataset the model is being tested on, OL = Offensive Language, HSB = Hate Speech Base, ED = Exstensive Definition, FoC = Form of Communication, PC, Problematic Content, T = Target).}
\label{tab:F1s_step1}
\end{table*}

\noindent
Table \ref{tab:F1s_step1} presents macro-F1 scores for different models and datasets, along with correlation values between performance and definition informativeness.

\begin{table*}[h]
\resizebox{\textwidth}{!}{
\begin{tabular}{@{}l|ccc|ccc|ccc!{}}
\toprule
\multicolumn{1}{c|}{\multirow{2}{*}{\textbf{Definitions}}} & \multicolumn{3}{c|}{\textbf{HateCheck}} & \multicolumn{3}{c|}{\textbf{LFTW}} & \multicolumn{3}{c}{\textbf{MHS}} \\
\cmidrule(lr){2-4} \cmidrule(lr){5-7} \cmidrule(lr){8-10}
 & LLama3 & Mistral & FlanT5 & LLama3 & Mistral & FlanT5 & LLama3 & Mistral & FlanT5 \\
\midrule
+LAA & 79.70 & 76.87 & 75.69 & \textbf{\underline{\underline{74.24}}} & \underline{56.05} & 63.04 & 73.12 & \underline{77.91} & 74.96 \\
+LAA\_PI & 77.39 & \underline{78.44} & \textbf{75.95} & 73.16 & \underline{\underline{58.40}} & \textbf{63.84} & 72.09 & \underline{77.71} & \textbf{75.30} \\
+LAA\_Exc & \textbf{79.72} & 75.95 & 75.30 & 72.96 & 54.22 & 62.31 & \textbf{\underline{74.67}} & 76.16 & 74.61 \\
+LAA\_IHS & 77.42 & \underline{\underline{80.74}} & 75.66 & 72.97 & \underline{\underline{60.97}} & 63.27 & 71.53 & \textbf{\underline{78.22}} & 74.91 \\
+LAA\_PI\_Exc & 76.65 & 73.88 & 75.38 & 73.22 & 53.00 & 62.99 & 73.60 & 75.56 & 75.14 \\
+LAA\_Exc\_IHS & 78.17 & \underline{78.27} & 75.76 & \underline{73.72} & \underline{56.38} & 63.34 & 74.55 & \underline{77.61} & 74.81 \\
+LAA\_PI\_IHS & 76.03 & \textbf{\underline{\underline{81.69}}} & 75.37 & 72.06 & \textbf{\underline{\underline{62.17}}} & 62.95 & 71.43 & \underline{78.06} & 74.48 \\
+LAA\_PI\_IHS\_Exc & 77.48 & \underline{\underline{78.62}} & 75.70 & 72.71 & \underline{\underline{57.92}} & 62.80 & 72.30 & \underline{77.88} & 75.00 \\
\bottomrule
\end{tabular}
}
\caption{Step 2, F1-macro: in \textbf{bold} the highest score in the step, the \underline{underlined} scores are those which are higher than the chosen crafted definitions of Step 1. Scores \underline{\underline{underlined twice}} are higher than the best performing definition of Step 1. (+ = best performing definition from Step 1, LAA = List of Addressed Attributes, PI = Possible Implications, Exc = Exception, IHS = Implicit Hate Speech).
}
\label{tab:F1s_step2}
\end{table*}

\noindent
\textbf{LLama-3} performs best without any definition (NO) in two out of three datasets, suggesting potential data leakage from HateCheck. While in the LFTW dataset, we encounter the only instance in which the best definition is the one of the dataset itself (Own). Among the crafted definitions, HSB + EDT performs best for HateCheck and LFTW, while HSB + EDFoC + EDPC is optimal for the MHS dataset.

\noindent
\textbf{Mistral} achieves its highest scores with either NO definition or Offensive Language (OL), implying an internalized concept of hate speech that aligns with offensive language. Among crafted definitions, HSB + EDFoC performs best in two datasets, while HSB is most effective in HateCheck.

\noindent
\textbf{Flan-T5}, unlike the other models, benefits consistently from definition prompting. Performance improves as definitions become more detailed, with HSB + EDT yielding the highest results in HateCheck and MHS, while the most extensive definition (HSB + EDFoC + EDPC + EDT) is optimal for LFTW.

%These findings highlight that models respond differently to definitional refinements. While \texttt{Flan-T5} benefits from increased informativeness, \texttt{Mistral} appears more sensitive to broad conceptual distinctions, and \texttt{LLama-3}’s performance is inconsistent, possibly due to pre-training influences.

\subsection{Step 2: Adding more Conceptual Elements to the optimal base definition}\label{subsec:step2}

Table \ref{tab:F1s_step2} presents the results of combining accessory elements with the best-crafted definition from step one.

\noindent
\textbf{LLama-3} improves in performance on the crafted definitions only in LFTW and Measuring Hate Speech, with the former surpassing the best performing definition (Own) of the previous step with +LAA.

\noindent
\textbf{Mistral} contrary to the previous step, is the most positively affected, improving its performance in different conditions over all the datasets concerning not only the crafted definitions. Reaching its new best performance in HateCheck with and LFTW both with +LAA + PI + IHS.

\noindent
\textbf{Flan-T5} shows an opposite trend compared to the previous step, where definition prompting has always led to an improvement in performance, here we do not observe in any condition a further increase in performance, though all the results are still higher than the condition without definition. 

Ultimately, we observe two consistent trends across the three datasets. \texttt{Mistral} improves only on the second step, when additional elements are added to the construct of HS, or in other words, some specificity of information is added to the definition. While \texttt{Flan-T5} shows improvement only in the first step, being thus more sensitive to the level of detail/informativeness of the definition, being also the only model which shows a positive correlation between performance and length of the definition (Table \ref{tab:F1s_step1}).

Performance-wise, we observe that while all models behave differently, their trends remain consistent across datasets. \texttt{LLama-3} generally does not show improvement, with a performance increase occurring only once on the LFTW dataset. In contrast, \texttt{Mistral} consistently improves in the second step, while \texttt{Flan-T5} shows gains in the first step, indicating that these models are more responsive to different types of information. \texttt{Mistral} benefits from more specific details, such as references to implicit HS, whereas \texttt{Flan-T5} responds to broader definitional refinements.

As a sanity check, we include additional analyses in the Appendix: robustness, which examines the stability of model performance across different runs (Appendix \ref{app:robustness}), and sensitivity, which measures how much model responses vary when different definitions are applied (Appendix \ref{app:sensitivity}).

\section{Error Analysis}

\paragraph{Hate Speech vs. Not Hate Speech}

\begin{figure*}[h]
    \centering
    \includegraphics[width=\textwidth]{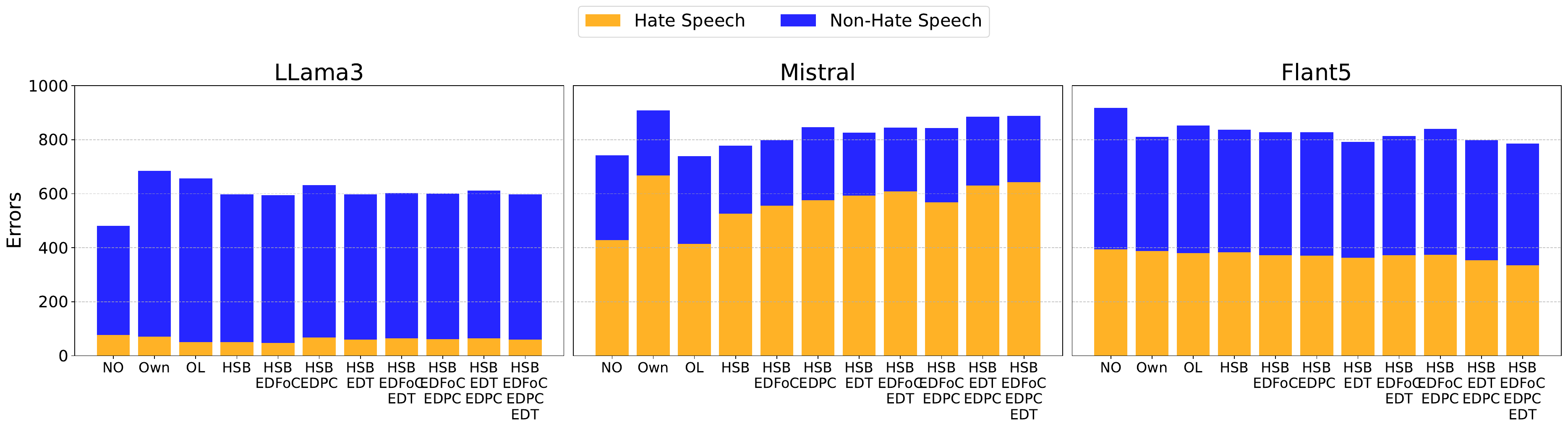}
    \caption{Distribution of errors across the three models on HateCheck.}
    \label{fig:Distr_err}
\end{figure*}

\begin{figure*}[h]
    \centering
    \includegraphics[width=\textwidth]{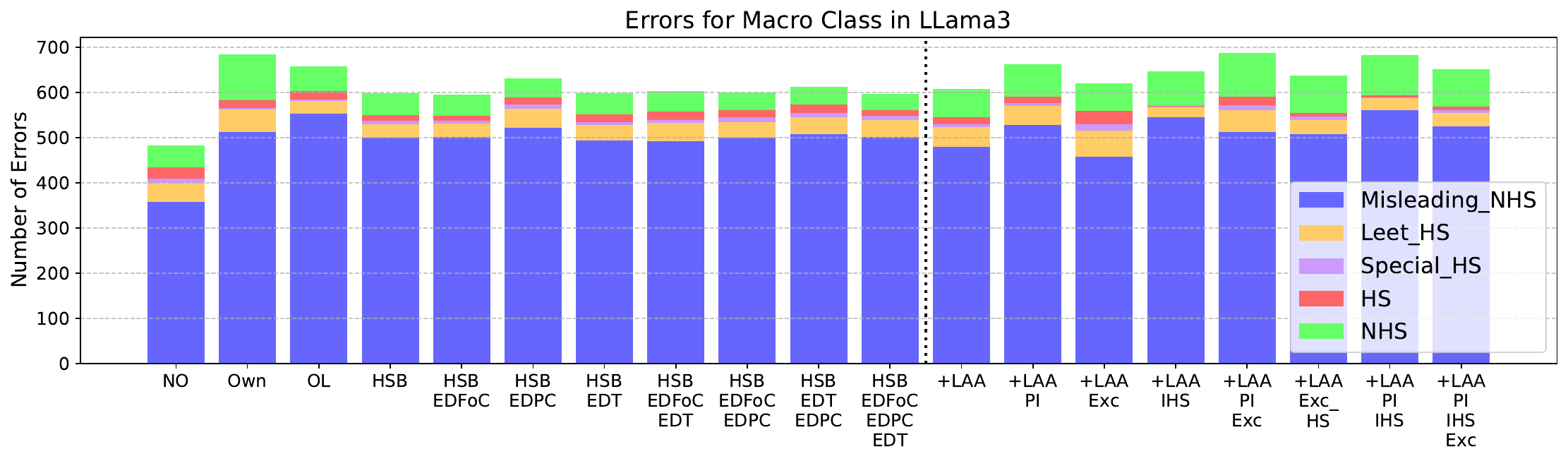}
    \caption{Distribution of errors in LLama-3 across Hatecheck Classes}
    \label{fig:LLama3_MCs}
\end{figure*}

A distinct model-dependent trend is evident in both the HS and NHS classes across all datasets. As shown in Fig. \ref{fig:Distr_err}, \texttt{LLama-3} frequently misclassifies NHS instances (i.e., non-hateful content) as HS, resulting in a higher false positive rate. This tendency appears to intensify when a definition is added to the prompt (e.g., changing from NO to Own). At the same time, introducing any definition reduces the number of misclassified HS instances (false negatives), suggesting that the model follows a more conservative classification approach.
\texttt{Mistral}, on the other hand, exhibits the opposite tendency, frequently misclassifying HS instances as NHS. However, adding a definition to the prompt reduces the number of false positives (misclassified NHS instances). \texttt{Flan-T5} maintains a more balanced classification pattern but shows a higher false negative rate, especially when tested on human-in-the-loop data points.\footnote{Appendix \ref{app:Error_anal_classes} contains detailed graphs of the error distribution across all conditions.}

\paragraph{Analysis of HateCheck functionalities}\label{subsec:type_of_hate}
HateCheck is a test suite for HS with a comprehensive labeling structure which defines each data point’s functionality— the specific type of hate conveyed. In this section, we conduct a micro-analysis on these HS functionalities to examine how classification performance changes across them when prompted with different CEs. The authors of HateCheck \cite{rottger2021hatecheck} identify 29 distinct functionalities. For easier comparability, we have grouped them into five macro classes: HS, NHS, Misleading NHS, Leet HS, and Special HS. Appendix \ref{app:functionalities} provides explanations and details on this grouping.

The error distribution across all functionalities \footnote{An overview of all errors are presented in Appendix \ref{app:functionalities}.} reveals that models struggle the most with instances designed to counter HS, such as statements like \textit{“If you say ‘I hate gay people,’ then you are a bigot!”}. Even the best-performing model, \texttt{LLama-3}, misclassifies 93.75\% of these cases, suggesting that LLMs still rely on specific words or sentence structures when making classifications.

Next, we examine how different definitions influence classification performance across the Macro Classes. Figure \ref{fig:LLama3_MCs} illustrates the variation in classification errors for \texttt{LLama-3} across all definitions. Overall, a more detailed definition tends to improve classification accuracy for general HS and NHS, with a slight positive effect on Leet HS and Special HS. However, it also leads to an increase in errors for the Misleading NHS class (Appendix \ref{app:MC_Graphs} presents results for all models). Furthermore, we find that adding a CE specific to a class of instances reduces classification errors for that class. For example, explicitly informing the model that some statements may be offensive but not hateful (i.e., defining exceptions) improves performance in the Misleading NHS class. A similar effect is observed for implicit hate speech: clarifying that hate speech can be conveyed through coded language, irony, or sarcasm leads to performance gains in the Special HS class. This effect is most pronounced in \texttt{Mistral}, the only model that consistently improves in the second step (see Table \ref{tab:F1s_step2}). We also observe a partial effect in \texttt{Flan-T5}, though it never improves in the second step \footnote{Appendix \ref{app:MC_Tables} provides detailed results for all models.}. Table \ref{tab:MC_type of hate_Main} presents a detailed breakdown of these effects. These findings have a potential relevance for content moderation, which we discuss in the conclusion.

\begin{table}[h]
    \resizebox{\columnwidth}{!}{
    \centering
    \begin{tabular}{lccccc}
        \toprule
        Mistral    & NO  & +IHS & +Exc & +Exc+IHS \\
        \midrule
        Misleading NHS  &  34.81\%  &  28.36\%  &   \textbf{23.26\%}  & 26.04\% \\
        Leet HS    &       20.81\%  &  \textbf{19.76\%}  &  29.23\%   & 24.24\% \\
        Special HS    &     18.75\%   &  \textbf{15.71\%} &  27.14\%   &  21.84\%  \\
        \bottomrule
    \end{tabular}
    }
    \caption{Error percentage, Conceptual Elements \& macro classes in Mistral.}
    \label{tab:MC_type of hate_Main}
\end{table}

%\textcolor{blue}{
%When we instead look at the definition prompted in the second step, we observe that when we add a CE which reflect a specific class of %data-points, the errors in that specific class decreases. In Table \ref{tab:MC_type of hate_Main} we provide breakout of these instances. 
%By looking at the first row, we see that the errors in the class Misleading NHS, those data-points which resemble HS but are not (see the counter HS example in the previous paragraph), decrease when when the CE of exception (Exc), which describes what is not considered HS, is added to the definition. However, when combined with other CEs, like the mention of implicit HS (which also leads to an improvement in classification in the respective macro classes), the performance seems to adjust in between, namely, providing smaller gains in performance, but in both the direction. 

%\vspace{-0.7cm}
\section{Conclusion}

In this work, we explored the conceptualization of the construct definition of HS and its influence on zero-shot prompting on three datasets. % including different data. %(synthetic, human-in-the-loop, real world examples). 
Our starting point has been the review of existing HS definitions, from which we inductively derived a set of Conceptual Elements. %, such as the target of hate, which can be expressed by a base definition, i.e., hate is directed towards "a group or an individual" (CE:Target) or in an extended way ""a group or an individual which is, or thought to be, a member of that group" (CE:Extensive Definition of Target). We identified 14 such elements, and arranged them in a taxonomy reflecting the varying degrees of informativeness (Extensive definitions being more informative than the Base ones). 
We then combined the different elements in the taxonomy to build a collection of definitions that lend themselves as prompts for LLM modeling.
Thus, the taxonomy and the collection definition are not just a conceptual contribution of our work, but also a concrete resource that can and should be used by researchers to structure their operationalization of the HS construct, thereby contributing to a clearer research landscape. 
Furthermore, the three-layers taxonomy, allows for combinations reflecting different levels of detail, which can be employed in annotation tasks in the descriptive vs. prescriptive paradigms \cite{rottger2021two}.

In our experiments, we exploited the definition collection for a series of zero-shot experiments, with the definitions serving as a series of curated prompts with increasing level of details.

%We utilize this new taxonomy in a zero-shot classification setting and find that choosing different construct definitions impacts model performance, but this effect is not consistent across all model architectures. For some models, like \texttt{Flan-T5}, we observe a clear trend where prediction performance increases with greater definition informativeness. However, in terms of performance, \texttt{Mistral} shows an opposite trend, improving only when specific elements are added to the base definition. While \texttt{LLama3} is the model which benefits from definition prompting only in one condition (Learning from the Worst, Second step). 

Our results show that varying construct definitions affects model performance, in a complex constellation of patterns. Some models benefit from detailed construct definitions by reducing false negatives, while others primarily decrease false positives. Our micro-analysis of different HateCheck functionalities shows that incorporating specific Conceptual Elements targeting particular types of hate improves model performance, especially in cases requiring nuanced distinctions between hate categories. 
Given our findings, we recommend that such a modular inspection of possible definitions should be employed for other complex constructs, beyond HS. 
Moreover, our findings do have practical implications for the usage of LLMs in production. Models that benefit from detailed construct definitions by reducing false negatives are, for example, better suited for high-recall moderation strategies, ensuring that fewer instances of hate speech go undetected. Conversely, models that primarily lower false positives are more appropriate for high-precision approaches, minimizing the risk of over-flagging benign content. By strategically refining definitions and incorporating targeted Conceptual Elements, moderation systems can be optimized to balance recall and precision according to platform-specific goals.

\section{Limitations}

Our study is not without limitations. A first one stems from computational restrictions. We were unable to test the largest model variants and assess their stability when prompted with different construct definitions. Furthermore, due to these computational constraints, we did not experiment with all possible construct definition combinations and settled on one fixed extensive definition. There is a possibility that different variants could have led to better performance.  

We also acknowledge that semantically different realization of the Conceptual Elements could have had a different impact on the models' performance. In other words surface-level phrasing, even when underlying CEs are held constant, can influence model behavior, an example of this can be seen on the MHS dataset, where the Own definition contains the exact same CEs of the HSB definition, though leading to different results. 

Another limitation is tied to the effect we have found in Sec. \ref{subsec:type_of_hate}. This is limited to the HateCheck datasets, to actually prove if this is a general effect, further studies should be conducted in annotated datasets. Our work only investigates performance differences in a zero-shot setting. It would be interesting to explore how carefully selected few-shot examples adhering to the given construct definition might impact stability and performance. Finally, we acknowledge that even the formulation of the prompt, without considering the construct definition itself, may influence the model's final performance.

\section{Ethical Considerations}

In our experiments, we do not collect any data, we instead use publicly available resources, so to ensure data protection. We also acknowledge that using Large Language Models to detect Hate Speech is not safe from issues, potentially not filtering appropriately and ending up spreading even more biases and discrimination. 
Furthermore, we made every effort to minimize content that could be disturbing or offensive, ensuring that any necessary reporting is handled appropriately and responsibly. 

\section{Acknowledgments}

The authors thank the funding received from the European Union’s Horizon Europe  Marie Skłodowska-Curie Actions Doctoral Networks, under Grant Agreement No. 101167978. This work was also conducted as part of the project Digital Dehumanization: Measurement, Exposure, and Prevalence (DeHum), supported by the Leibniz Association Competition (P101/2020).

\bibliography{anthology,custom} 

\appendix

\onecolumn
\section{Collection of Hate Speech definitions}\label{app:definitions}

\begin{longtable*}[h]{p{3cm}p{10cm}}
\toprule
\textbf{Authors} & \textbf{Definition} \\
\midrule
\citealp{nockleby2000hate} & Any communication that disparages a person or a group on the basis of some characteristic such as race, color, ethnicity, gender, sexual orientation, nationality, religion, or other characteristic. \\
\hline
\citealp{tsesis2002destructive} & Hate speech provides the “vocabulary and grammar depicting a common enemy,” and establishes a “mutual interest in trying to rid society of the designated pest.” \\
\hline
\citealp{warner2012detecting} & Hate speech is
defined as abusive speech targeting specific
group characteristics, such as ethnic origin, religion, gender, or sexual orientation. \\
\hline
\citealp{nobata2016abusive} & An act that attacks or demeans a group/individual based on race, ethnic origin, religion, disability, gender, age, disability, or sexual orientation/gender identity. \\
\hline
\citealp{Davidson} & Hate speech is language that is used to express hatred towards a targeted group or is intended to be derogatory, to humiliate, or to insult the members of the group. \\
\hline
\citealp{gao2017recognizing} & Hateful speech is defined as the language which explicitly or implicitly threatens or demeans a person or a group based upon a facet of their identity such as gender, ethnicity, or sexual orientation. \\
\hline
\citealp{founta2018large} & Hate speech is language used to express hatred towards a targeted individual or group, or is intended to be derogatory, to humiliate, or to insult the members of the group, on the basis of attributes such as race, religion, ethnic origin, sexual orientation, disability, or gender. \\
\hline
\citealp{gibert2018hate} & Hate speech is any communication that disparages a target group of people based on some characteristic such as race, color, ethnicity, gender, sexual orientation, nationality, religion, or other characteristic. \\
\hline
\citealp{basile2019semeval} & Hate speech is commonly defined as any communication that disparages a person or a group on the basis of some characteristic such as race, color, ethnicity, gender, sexual orientation, nationality, religion, or other characteristics. \\
\hline
\citealp{mandl2021overview} & Hate speech includes ascribing negative attributes or deficiencies to groups of individuals because they are members of a group (e.g. all poor people are stupid). This class combines any hateful comments toward groups because of race, political opinion, sexual orientation, gender, social status, health condition, or similar. \\
\hline
\citealp{rottger2021hatecheck} & Hate speech is abuse that is targeted at a protected group or at its members for being a part of that group. Protected groups are defined based on age, disability, gender identity, familial status, pregnancy, race, national or ethnic origins, religion, sex or sexual orientation, which broadly reflects international legal consensus (particularly the UK’s 2010 Equality Act, the US 1964 Civil Rights Act, and the EU’s Charter of Fundamental Rights). \\
\hline
\citealp{vidgen-etal-2021-learning} & Hate is defined as abusive speech targeting specific group characteristics, such as ethnic origin, religion, gender, or sexual orientation. \\
\hline
\citealp{sachdeva2022measuring} & Hate speech, defined as "bias-motivated, hostile and malicious language targeted at a person/group because of their actual or perceived innate characteristics, especially when the group is unnecessarily labeled" \\
\hline
UN Strategy and Plan of Action on Hate Speech & Any kind of communication in speech, writing or behavior, that attacks or uses pejorative or discriminatory language with reference to a person or a group on the basis of who they are, in other words, based on their religion, ethnicity, nationality, race, color, descent, gender or other identity factor. \\
\hline
Code of Conduct between European Union Commission and companies & All conduct publicly inciting to violence or hatred directed against a group of persons or a member of such a group defined by reference to race, color, religion, descent or national or ethnic. \\
\hline
\href{https://chat.openai.com/}{ChatGPT}'s definition & Hate speech typically refers to any form of communication – whether spoken, written, or expressed through actions – that seeks to demean, intimidate, discriminate against, or incite violence or prejudice against individuals or groups based on characteristics such as race, ethnicity, nationality, religion, gender identity, sexual orientation, disability, or any other immutable characteristic. \\
\hline
\href{https://chat.openai.com/}{Gemini}'s definition & Hate speech is basically language that attacks a person or group based on things they can't control, like their race, religion, gender, sexual orientation, or disability. \\
\hline
\href{https://www.facebook.com}{Facebook} & We define hate speech as a direct attack against people on the basis of what we call protected characteristics: race, ethnicity, national origin, disability, religious affiliation, caste, sexual orientation, sex, gender identity, and serious disease. \\
\hline
\href{https://twitter.com/}{Twitter/X}  & You may not promote violence against, threaten, or harass other people on the basis of race, ethnicity, national origin, caste, sexual orientation, gender, gender identity, religious affiliation, age, disability, or serious disease. \\
\hline
\href{https://www.youtube.com/}{Youtube} & We remove content promoting violence or hatred against individuals or groups based on any of the following attributes: age, caste, disability, ethnicity, gender identity and expression, nationality, race, immigration status, religion, sex/gender, sexual orientation, victims of a major violent event and their kin, and veteran Status. \\
\bottomrule
\end{longtable*}

\section{Taxonomy: visual representation}\label{app:taxi}

\begin{figure}[H]
    \centering
    \includegraphics[width=\textwidth]{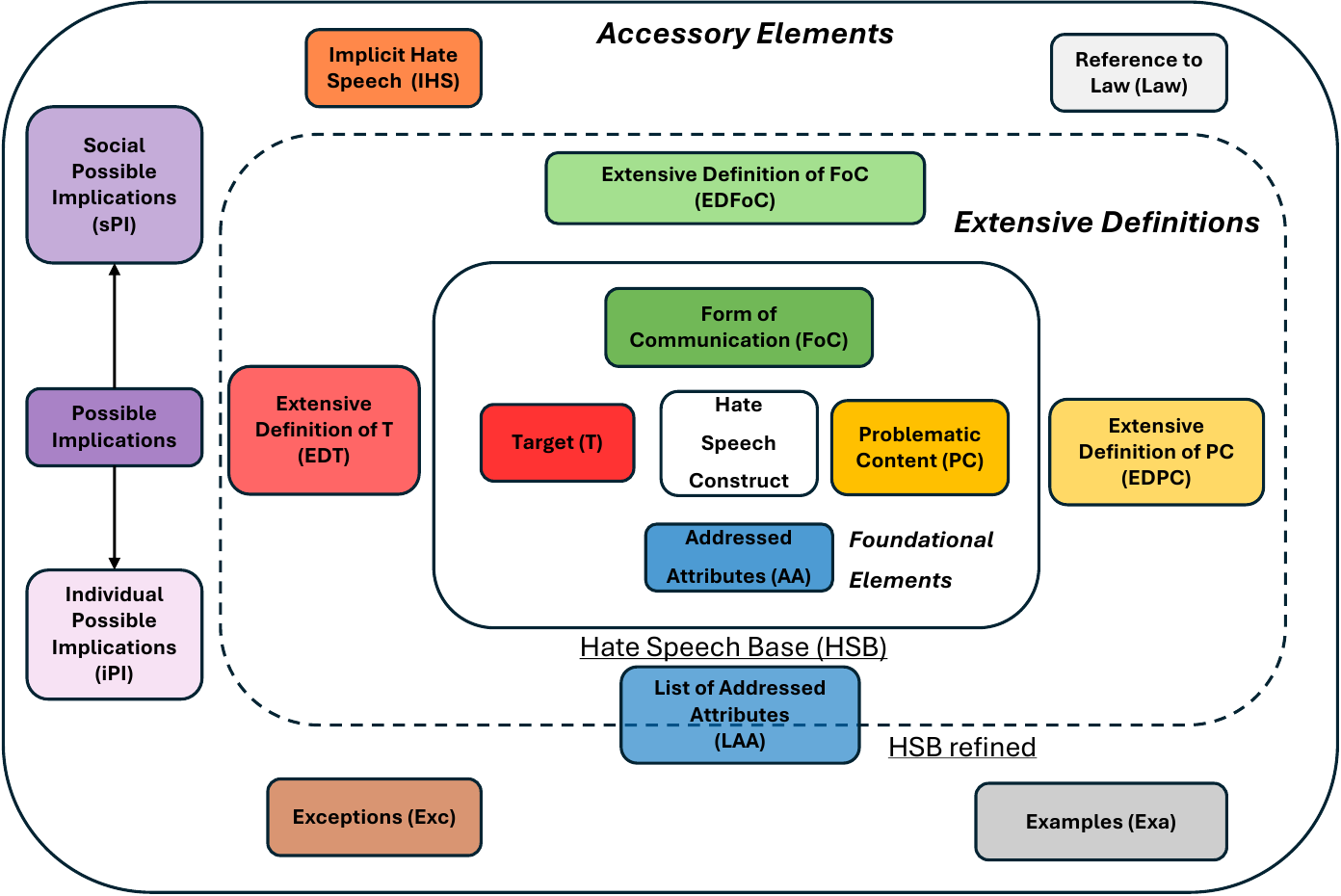}
    \caption{Graphic visualization of the taxonomy.}
    \label{fig:taxi}
\end{figure}

\section{Conceptual Elements in our Taxonomy: Definitions}\label{app:CE_def}

\textbf{FOUNDATIONAL ELEMENTS:}
We define Foundational Elements as those Conceptual Elements that are required to build a meaningful definition of Hate Speech. In this category we find:
\newline

\noindent
\textbf{Form of Communication (FoC):} represents how the message is expressed. Refers to the modality of transmission, it can be text, speech, image, or video, … This element is independent from all the others, it is what grounds the HS to the real world and it is the whole “container” of the HS. 
\newline

\noindent
\textbf{Target (T):} represents toward whom the FoC is directed (individual or group). It describes the real word entity that is addressed by the FoC. We can identify it as the object of the message. In the FoC it is often identified as a social category (”black people are…”), a slur that identifies a member of that category or the category itself (”n-word”).
\newline

\noindent
\textbf{Problematic Content (PC):} represents realisation of the malevolent communicative intent conveyed by a specific FoC. It describes that part of the FoC that has a negative connotation and it is implied to be a derogatory descriptor of T. It is the form (in our case, linguistics) in which the malevolent communicative intent is expressed in the FoC. It can be a sentiment (”I hate..”, “I can’t bear…”) a slur (”gay people are all dumb”), or anything that implies negativity toward the T.
\newline

\noindent
PC and T can assume multiple forms and sometimes overlap (e.g., n-word, f-word), and they are both dependent form the FoC — without it there cannot be PC and T. 
\newline

\noindent
\textbf{Addressed Attributes (AA):} represents that part of the FoC that is specific to the Hate Speech and explicitly describes the relation between PC and T. In other words it describes which are the aspects of the T that motivate the malevolent communicative intent and thus the creation of a PC. It describes that the malevolent communicative intent has to specifically aim to a group or a person that belongs to a group and to the inherent characteristics that the group and the individual share or are thought to share. Thought being part of the definition, it can also be found in HS comment: being  explicitly expressed in the FoC (”I hate black people [skin color]”) or take the form of a generalisation (”[All] disabled people are stupid”) or can be left implicit overlapping with the other elements (for instance i with the Target: ”[Affirmative action] means we get affirmatively second rate doctors and other professionals”).
\newline

\noindent
Being these the Foundational Elements of the construct definition of Hate Speech, different combination of them will lead to constructs different than HS, here below we provide four examples of different combination.

\begin{enumerate}
 
    \item If PC and AA are missing, the communication (FoC, T) is not Hate Speech, but it is just communication. 

    \item If T and AA are missing, the communication (FoC, PC)  \textbf{can} still be offensive (or toxic), but not categorized as Hate Speech  (i.e., “this is bul***it”,  “Cauliflowers are fu**ing disgusting”).

    \item If AA is missing, the communication (FoC, T, PC) it is not Hate Speech but Offensive Language (“[POLITICIAN NAME] is the dumbest politician in the US"). 
    
    \item There are no cases in which there are only PC (FoC, PC, AA) and AA or T and AA (FoC, T, AA). This makes AA dependent from PC and T (other than from the FoC). It comes that, when it seems to have a case of this kind, actually AA overlaps with the apparent “missing Conceptual Element" (i.e., AA overlaps with PC “you are a [f-word]”).
    
\end{enumerate}

\noindent
\textbf{EXTENSIVE DEFINITIONS:} 
Are those elements that provide further information about the construct, and can be used go implement further levels of details/informativeness of the construct definition. First we have identified a group of Conceptual Elements, Extensive Definitions (EDs) that add further information to the Foundational elements, In other words they do not provide different pieces information from those already in the definition (HSB), but only describe more in details the pieces of information provided by the Foundational Elements.
\newline.

\noindent
\textbf{Extensive Definition Form of Communication (EDFoC):} other ways to describe the FoC, in our case, it is important that it is explained as text or language or communication, however, this Accessory element provides another way in which the Hate Speech can be transmitted (e.g.,“Hate speech can manifest in various forms including but not limited to verbal attacks","any form of communication – whether spoken, written, or expressed through actions").
\newline

\noindent
\textbf{Extensive Definition Target (EDT):} specifies the relation between a person and being a member of a group, the idea of “belonging to a group”.
\newline

\noindent
\textbf{Extensive Definition Problematic Content (EDPC):} it gives more information and better describes PC, providing examples of what is considered PC; (i.e., ”We define attacks as violent or dehumanizing speech, harmful stereotypes, statements of inferiority, expressions of contempt, disgust or dismissal, cursing, and calls for exclusion or segregation.”).
\newline

\noindent
\textbf{List of Addressed Attributes (LAA):}  it provides a list of characteristics/attributes of the T that can be object of the PC (i.e., "such as race, gender, religion, ..").
\newline

\noindent
\textbf{ACCESSORY ELEMENTS:} 
Finally, we define as Accessory Elements those elements that provide different information on the construct of HS, namely, information that it is not present in the HSB definition and describes other aspects of the HS construct. In
\newline

\noindent
\textbf{Possible Implications (PI):} part of the FoC that refers to the possible consequences of a particular combination of PC and T. It can be divided into two sublevels:
\begin{enumerate}
    \item \textit{social (sPI):} it refers to the implication on the social level of one (or more) PC toward a T (i.e., “undermines social cohesion, promotes division … in communities”).
    \item \textit{individual (iPI):} it refers directly to the effects that one (or more) PC can have on the T (i.e., “can have serious consequences for individuals, often perpetuating discrimination, hostility, and violence”).
\end{enumerate}

\noindent
\textbf{Exceptions (Exc):} provide information on what is not considered HS (i.e., ”attacks on people's personality traits, ideas, or opinions”).
\newline

\noindent
\textbf{Implicit Hate Speech (IHS):} Hate speech is not always explicit, this conceptual element describes what is considered Implicit Hate Speech, conceptually a communication that is missing a conceptual element among Target, Problematic Content and Addressed Attributes. To define this conceptual element we have been inspired by \citealp{ghosh2023cosyn} and \citealp{elsherief2021latent}.
\newline

\noindent
The following two Conceptual Elements were not implemented in our experiment. The first in order to maintain a Zero-Shot-Learning condition, while the second would have implied an to investigate if the models actually knows the laws that we are referring to, and this was not in the scope of our research.
\newline

\noindent
\textbf{Examples (Exa):} the information provided by this CE is simply an instance of a sentence that it is considered Hate Speech.
\newline

\noindent
\textbf{Reference to laws: (Law):} part of the definition that provide information in regards to specific laws that regulate Hate Speech. 
\newpage

\section{Conceptual Elements in HS the literature: overview }\label{app:CE}
\begin{table}[h!]
\centering 
\small
\rotatebox{90}{\begin{tabular}{l|cccccccccccccc}
    \toprule
    \textbf{Author} & \textbf{FoC} & \textbf{T} & \textbf{PC} & \textbf{AA} & \textbf{EDFoC} & \textbf{EDT} & \textbf{EDPC} & \textbf{LAA} & \textbf{sPI} & \textbf{iPI} & \textbf{Exc} & \textbf{IHS} & \textbf{Exa} & \textbf{Law} \\
    \midrule
    \citealp{nockleby2000hate} & \cmark & \cmark & \cmark & \xmark & \xmark & \xmark & \xmark & \cmark & \xmark & \xmark & \xmark & \xmark & \xmark & \xmark \\
    \citealp{tsesis2002destructive} & \cmark & ? &  \cmark & \xmark & \xmark & \xmark & \xmark & \xmark & \xmark & \xmark & \xmark & \xmark & \xmark & \xmark \\
    \citealp{warner2012detecting} & \cmark & \cmark & \cmark & \xmark & \xmark & \xmark & \xmark & \cmark & \xmark & \xmark & \xmark & \xmark & \xmark & \xmark \\
    \citealp{nobata2016abusive} & ? & \cmark & \cmark & \xmark & \xmark & \xmark & \xmark & \cmark & \xmark & \xmark & \xmark & \xmark & \xmark & \xmark \\
    \citealp{Davidson} & \cmark & \cmark & \cmark & \xmark & \xmark & \xmark & \cmark & \xmark & \xmark & \xmark & \xmark & \xmark & \xmark & \xmark \\
    \citealp{gao2017recognizing} & \cmark & \cmark & \cmark & \cmark & \xmark & \xmark & \xmark & \cmark & \xmark & \xmark & \xmark & \xmark & \xmark & \xmark \\
    \citealp{founta2018large}& \cmark & \cmark & \cmark & \xmark & \xmark & \cmark & \cmark & \xmark & \xmark & \xmark & \xmark & \xmark & \xmark & \xmark \\
    \citealp{gibert2018hate}& \cmark & \cmark & \cmark & \xmark & \xmark & \xmark & \xmark & \cmark & \xmark & \xmark & \xmark & \xmark & \xmark & \xmark \\
    \citealp{basile2019semeval}& \cmark & \cmark & \cmark & \xmark & \xmark & \xmark & \xmark & \cmark & \xmark & \xmark & \xmark & \xmark & \xmark & \xmark \\
    \citealp{mandl2021overview} & \cmark & \cmark & \cmark & \cmark & \xmark & \xmark & \xmark & \cmark & \xmark & \xmark & \xmark & \xmark & \cmark & \xmark \\
    \citealp{rottger2021hatecheck} & \cmark & \cmark & \cmark & \cmark & \xmark & \xmark & \xmark & \cmark & \xmark & \xmark & \xmark & \xmark & \xmark & \cmark \\
    \citealp{vidgen-etal-2021-learning} & \cmark & \cmark & \cmark & \cmark & \xmark & \xmark & \xmark & \cmark & \xmark & \xmark & \xmark & \xmark & \xmark & \xmark \\
    \citealp{sachdeva2022measuring} & \cmark & \cmark & \cmark & \cmark & \xmark & \cmark & \xmark & \xmark & \xmark & \xmark & \xmark & \xmark & \xmark & \xmark \\
    \href{https://twitter.com/}{Twitter/X} & \xmark & \cmark & \cmark & \xmark & \xmark & \xmark & \xmark & \cmark & \xmark & \xmark & \xmark & \xmark & \cmark & \xmark \\
    \href{https://www.facebook.com}{Facebook}  & \cmark & \cmark & \cmark & \xmark & \xmark & \xmark & \cmark & \cmark & \xmark & \xmark & \cmark & \xmark & \xmark & \xmark \\
    \href{https://www.youtube.com/}{YouTube}  & \cmark & \cmark & \cmark & \xmark & \xmark & \xmark & \xmark & \cmark & \xmark & \xmark & \xmark & \xmark & \xmark & \xmark \\
    \href{https://chat.openai.com/}{ChatGPT}  & \cmark & \cmark & \cmark & \xmark & \cmark &  \xmark & \cmark & \cmark & \cmark & \cmark & \xmark & \xmark & \xmark & \xmark \\
    \href{https://chat.openai.com/}{Gemini}  & \cmark & \cmark & \cmark & \cmark & \cmark & \xmark & \cmark & \cmark & \cmark & \xmark & \xmark & \cmark & \xmark & \cmark \\
    \href{https://www.un.org/en/hate-speech/understanding-hate-speech/what-is-hate-speech}{UN Strategy \& Plan of Action on HS}  & \cmark & \cmark & \cmark & \cmark & \cmark & \xmark & \xmark & \cmark & \xmark & \xmark & \xmark & \xmark & \xmark & \xmark \\
    Code of Conduct EU (\citealp{wigand2017speech})  & \cmark  & \cmark  & \cmark  & \xmark & \cmark  & \cmark & \xmark & \cmark & \xmark & \xmark & \xmark & \xmark & \xmark & \xmark \\
    \bottomrule
  \end{tabular}}
  \captionsetup{justification=centering}
  \caption{Outline of the Conceptual Elements in the collected definitions.   \\
  \cmark = present in the definition, \xmark = absent, ? = present but we consider it too vague to be part of a definition.}
\end{table}
\clearpage

\section{Collection of Definition Prompted in the Experiment}\label{app:definition_prompted}

\begin{longtable}{p{\linewidth}} 
    \toprule
    \textbf{OL - Offensive Language} \\ 
    \midrule
    Hate Speech is considered any kind of content that conveys malevolent intentions toward a group or an individual. \\
    \midrule
    \endfirsthead

    \toprule
    \textbf{(Continued) Definitions of Offensive and Hate Speech} \\
    \midrule
    \endhead

    \midrule
    \multicolumn{1}{r}{{Continued on next page}} \\ 
    \midrule
    \endfoot

    \bottomrule
    \endlastfoot

    \textbf{HSB - Hate Speech Base} \\
    \midrule
    Hate speech is considered any kind of content that conveys malevolent intentions toward a group or an individual, and is motivated by inherent characteristics that are attributed to that group and shared among its members. \\
    \midrule
    \textbf{HSB\_EDFoC - Hate Speech Base + Extensive Definitions of Form of Communication} \\
    \midrule
    Hate speech is considered any kind of content or communication expressed using language (written or spoken) or actions, that conveys malevolent intentions toward a group or an individual, and is motivated by inherent characteristics that are attributed to that group and shared among its members. \\
    \midrule
    \textbf{HSB\_EDPC - Hate Speech Base + Extensive Definitions of Problematic Content} \\
    \midrule
    Hate speech is considered any kind of content that conveys malevolent intentions such as statements of inferiority, aversion, cursing, calls for exclusion, threats, harassment, or violence, toward a group or an individual, and is motivated by inherent characteristics that are attributed to that group and shared among its members. \\
    \midrule
    \textbf{HSB\_EDT - Hate Speech Base + Extensive Definitions of Target} \\
    \midrule
    Hate speech is considered any kind of content that conveys malevolent intentions toward a group or an individual who is, or is thought to be, a member of that group, and is motivated by inherent characteristics that are attributed to that group and shared among its members. \\
    \midrule
    \textbf{HSB\_EDFoC\_EDT - Hate Speech Base + Extensive Definitions of: Form of Communication and Target} \\
    \midrule
    Hate speech is considered any kind of content or communication expressed using language (written or spoken) or actions, that conveys malevolent intentions toward a group or an individual who is, or is thought to be, a member of that group, and is motivated by inherent characteristics that are attributed to that group and shared among its members. \\
    \midrule
    \textbf{HSB\_EDFoC\_EDPC - Hate Speech Base + Extensive Definitions of: Form of Communication and Problematic Content} \\
    \midrule
    Hate speech is considered any kind of content or communication expressed using language (written or spoken) or actions, that conveys malevolent intentions such as statements of inferiority, aversion, cursing, calls for exclusion, threats, harassment, or violence, toward a group or an individual, and is motivated by inherent characteristics that are attributed to that group and shared among its members. \\
    \midrule
    \textbf{HSB\_EDT\_EDPC - Hate Speech Base + Extensive Definitions of: Problematic Content and Target} \\
    \midrule
    Hate speech is considered any kind of content that conveys malevolent intentions such as statements of inferiority, aversion, cursing, calls for exclusion, threats, harassment, or violence, toward a group or an individual who is, or is thought to be, a member of that group, and is motivated by inherent characteristics that are attributed to that group and shared among its members. \\
    \midrule
    \textbf{HSB\_EDFoC\_EDPC\_EDT - Hate Speech Base + Extensive Definitions of: Form of Communication, Problematic Content, and Target} \\
    \midrule
    Hate speech is considered any kind of content or communication expressed using language (written or spoken) or actions, that conveys malevolent intentions such as statements of inferiority, aversion, cursing, calls for exclusion, threats, harassment, or violence, toward a group or an individual who is, or is thought to be, a member of that group, and is motivated by inherent characteristics that are attributed to that group and shared among its members. \\
    \midrule
    \midrule
    \textit{For the second step, we take the HSB (Hate Speech Base) definition as exemplification and we add the combinations of the Accessory Elements.} \\
    \midrule
    \midrule
    \textbf{HSB\_LAA - Hate Speech Base + List of Addressed Attributes} \\
\midrule
Hate speech is considered any kind of content that conveys malevolent intentions toward a group or an individual, and motivated by inherent characteristics that are attributed to that group and shared among its members such as race, color, ethnicity, gender, sexual orientation, nationality, religion, disability, social status, health conditions, or other characteristics. \\
\midrule
\textbf{HSB\_LAA\_PI - Hate Speech Base + List of Addressed Attributes + Possible Implications} \\
\midrule
Hate speech is considered any kind of content that conveys malevolent intentions toward a group or an individual, and motivated by inherent characteristics that are attributed to that group and shared among its members such as race, color, ethnicity, gender, sexual orientation, nationality, religion, disability, social status, health conditions, or other characteristics. The outcome of Hate Speech could be the promotion of division among people, undermining of social cohesion in communities, inciting others to commit violence or discrimination, and could have consequences for individuals’ health and safety. \\
\midrule
\textbf{HSB\_LAA\_Exc - Hate Speech Base + List of Addressed Attributes + Exception} \\
\midrule
Hate speech is considered any kind of content that conveys malevolent intentions toward a group or an individual, and motivated by inherent characteristics that are attributed to that group and shared among its members such as race, color, ethnicity, gender, sexual orientation, nationality, religion, disability, social status, health conditions, or other characteristics. However, even if it is offensive, it is not considered Hate Speech any content that attacks a person’s personality traits, ideas, or opinions. \\
\midrule
\textbf{HSB\_LAA\_IHS - Hate Speech Base + List of Addressed Attributes + Implicit Hate Speech} \\
\midrule
Hate speech is considered any kind of content that conveys malevolent intentions toward a group or an individual and motivated by inherent characteristics that are attributed to that group and shared among its members such as race, color, ethnicity, gender, sexual orientation, nationality, religion, disability, social status, health conditions, or other characteristics. Hate Speech can also be implicit, portrayed as an indirect or coded language that uses Irony, Stereotypes, or Misinformation. \\
\midrule
\textbf{HSB\_LAA\_PI\_Exc -Hate Speech Base + List of Addressed Attributes + Possible Implication + Exceptions} \\
\midrule
Hate speech is considered any kind of content that conveys malevolent intentions toward a group or an individual, and motivated by inherent characteristics that are attributed to that group and shared among its members such as race, color, ethnicity, gender, sexual orientation, nationality, religion, disability, social status, health conditions, or other characteristics. The outcome of Hate Speech could be the promotion of division among people, undermining of social cohesion in communities, inciting others to commit violence or discrimination, and could have consequences for individuals’ health and safety. However, even if it is offensive, it is not considered Hate Speech any content that attacks a person’s personality traits, ideas, or opinions. \\
\midrule
\textbf{HSB\_LAA\_Exc\_IHS - Hate Speech Base + List of Addressed Attributes + + Exceptions + Implicit Hate Speech} \\
\midrule
Hate speech is considered any kind of content that conveys malevolent intentions toward a group or an individual, and motivated by inherent characteristics that are attributed to that group and shared among its members such as race, color, ethnicity, gender, sexual orientation, nationality, religion, disability, social status, health conditions, or other characteristics. Hate Speech can also be implicit, portrayed as an indirect or coded language that uses Irony, Stereotypes, or Misinformation. However, even if it is offensive, it is not considered Hate Speech any content that attacks a person’s personality traits, ideas, or opinions. \\
\midrule
\textbf{HSB\_LAA\_PI\_IHS - Hate Speech Base + List of Addressed Attributes + Possible Implications + Implicit Hate Speech} \\
\midrule
Hate speech is considered any kind of content that conveys malevolent intentions toward a group or an individual, and motivated by inherent characteristics that are attributed to that group and shared among its members such as race, color, ethnicity, gender, sexual orientation, nationality, religion, disability, social status, health conditions, or other characteristics. The outcome of Hate Speech could be the promotion of division among people, undermining of social cohesion in communities, inciting others to commit violence or discrimination, and could have consequences for individuals’ health and safety. Hate Speech can also be implicit, portrayed as an indirect or coded language that uses Irony, Stereotypes, or Misinformation. \\
\midrule
\textbf{HSB\_LAA\_PI\_IHS\_Exc - Hate Speech Base + List of Addressed Attributes + Possible Implications + Implicit Hate Speech + Exception} \\
\midrule
Hate speech is considered any kind of content that conveys malevolent intentions toward a group or an individual, and motivated by inherent characteristics that are attributed to that group and shared among its members such as race, color, ethnicity, gender, sexual orientation, nationality, religion, disability, social status, health conditions, or other characteristics. The outcome of Hate Speech could be the promotion of division among people, undermining of social cohesion in communities, inciting others to commit violence or discrimination, and could have consequences for individuals’ health and safety. Hate Speech can also be implicit, portrayed as an indirect or coded language that uses Irony, Stereotypes, or Misinformation. However, even if it is offensive, it is not considered Hate Speech any content that attacks a person’s personality traits, ideas, or opinions. \\
\bottomrule
\caption{Collection of definition prompted.} \\

\end{longtable}

\section{Prompt Templates}\label{app:prompts}

\begin{table*}[h]
    \centering
    \begin{tabular}{p{7cm}|p{7cm}}
    \toprule
    \textbf{Without definition} & \textbf{With definition} \\ \midrule
    \textit{Classify if the following text is considered hate speech or not hate speech:\textbackslash n
    [TEXT]\textbackslash n
    Please answer ONLY with 1 if it is hate speech or 0 if it is not hate speech.} &
    \textit{Given the following definition of Hate Speech: [Definition]\textbackslash n
    Classify if the following text is considered hate speech or not hate speech:\textbackslash n
    [TEXT]\textbackslash n
    Please answer ONLY with 1 if it is hate speech or 0 if it is not hate speech.} \\
    \bottomrule
    \end{tabular}
    \caption{Prompts used for the classification task}
    \label{tab:prompts}
\end{table*}

\newpage
\section{Robustness}\label{app:robustness}

We measure robustness by checking how many times the models answer in the same way under the three runs. All the results are reported below in Table \ref{tab:const_within}. In general, we observe a high consistency which stays relatively stable across models and datasets. Given the overall similarity between the scores, we identified outliers using the Interquartile Range (IQR), we observe that on the first step, 6 out of 9 times the outliers are the scores obtained in the condition without definition (NO). While on the second step, the definitions which contain the CE of exception generally lead to less robustness. Finally, we do not observe particular trends related to the highest value obtained in either of the steps. 

\begin{table*}[h]
\resizebox{\textwidth}{!}{
\captionsize
\begin{tabular}{@{}l|ccc|ccc|ccc!{}}
\toprule
\multicolumn{1}{c|}{\multirow{2}{*}{\textbf{Definitions}}} & \multicolumn{3}{c|}{\textbf{HateCheck}} & \multicolumn{3}{c|}{\textbf{Learning from the Worst}} & \multicolumn{3}{c}{\textbf{Measuring Hate Speech}} \\
\cmidrule(lr){2-4} \cmidrule(lr){5-7} \cmidrule(lr){8-10}
 & LLama3 & Mistral & FlanT5 & LLama3 & Mistral & FlanT5 & LLama3 & Mistral & FlanT5 \\
\midrule
NO & \underline{91.28\%} & 96.00\% & \underline{70.34\%} & \underline{82.26\%} & \underline{93.92\%} & \underline{57.88\%} & 88.75\% & 95.87\% & \underline{72.78\%} \\
Own & 94.03\% & 96.46\% & 75.44\% & 86.77\% & 95.95\% & 65.34\% & 89.64\% & 96.15\% & 76.95\% \\
OL & 94.44\% & \textbf{96.87\%} & 75.11\% & 87.85\% & 94.59\% & 64.24\% & \underline{85.82\%} & \underline{94.69\%} & 75.85\% \\
HSB & 93.95\% & 96.51\% & 74.67\% & 88.29\% & 95.31\% & 65.65\% & 88.57\% & 96.46\% & 76.31\% \\
HSB\_EDFoC & 95.08\% & 96.49\% & 75.67\% & 88.29\% & 95.33\% & 66.91\% & 90.11\% & 96.39\% & 77.26\% \\
HSB\_EDPC & 94.77\% & 96.41\% & 74.06\% & 87.46\% & \textbf{95.98\%} & 65.62\% & 89.11\% & \textbf{96.69\%} & 77.39\% \\
HSB\_EDT & 94.26\% & 96.56\% & \textbf{76.83\%} & 88.23\% & 95.67\% & 67.03\% & 89.39\% & 96.23\% & 76.72\% \\
HSB\_EDFoC\_EDT & 94.23\% & 95.69\% & 76.06\% & \textbf{88.75\%} & 95.57\% & 65.24\% & 90.16\% & 95.85\% & 77.77\% \\
HSB\_EDFoC\_EDPC & 95.21\% & 96.18\% & 74.44\% & 87.23\% & 95.49\% & 66.50\% & 88.95\% & 96.44\% & \textbf{78.19\%} \\
HSB\_EDT\_EDPC & 94.77\% & 95.95\% & 75.85\% & 87.26\% & 95.85\% & \textbf{67.55\%} & 89.03\% & 96.54\% & 76.16\% \\
HSB\_EDFoC\_EDPC\_EDT & \textbf{95.36\%} & 96.33\% & 76.65\% & 87.23\% & 95.72\% & 67.50\% & \textbf{90.59\%} & 96.23\% & 78.11\% \\
\midrule
\midrule
Avg. Step 1 & \textbf{94.31\%} & 96.31\% & \textbf{75.01\%} & \textbf{87.23\%} & 95.39\% & 65.40\% & 89.10\% & 96.14\% & \textbf{76.68\%} \\
\bottomrule
\toprule
+LAA & 94.23\% & 96.41\% & \underline{\textbf{76.19\%}} & 87.64\% & 96.08\% & 66.21\% & 90.41\% & 96.64\% & \textbf{76.85\%} \\
+LAA\_PI & 94.46\% & 96.82\% & 74.47\% & 88.21\% & \textbf{96.28\%} & \textbf{67.91\%} & 92.26\% & \textbf{97.36\%} & 76.29\% \\
+LAA\_Exc & \underline{92.03\%} & 96.62\% & 74.31\% & \underline{82.62\%} & 95.39\% & 64.47\% & \underline{86.29\%} & 96.44\% & 74.26\% \\
+LAA\_IHS & 95.16\% & 97.23\% & 73.93\% & 88.13\% & 95.39\% & 65.42\% & 90.75\% & 96.69\% & 74.96\% \\
+LAA\_PI\_Exc & 92.54\% & 97.03\% & 74.72\% & \underline{84.54\%} & 96.15\% & 66.78\% & 88.39\% & 96.23\% & 76.72\% \\
+LAA\_Exc\_IHS & 94.05\% & 97.15\% & 74.31\% & 87.16\% & 95.44\% & 67.14\% & 90.39\% & 96.46\% & 74.08\% \\
+LAA\_PI\_IHS & \textbf{95.49\%} & \textbf{97.46\%} & 74.37\% & \textbf{88.31\%} & 95.59\% & 65.34\% & \textbf{92.44\%} & 97.15\% & 75.70\% \\
+LAA\_PI\_IHS\_Exc & 94.41\% & 96.80\% & 75.06\% & 87.11\% & 96.23\% & 66.42\% & 90.57\% & 96.56\% & 75.19\% \\
\midrule
\midrule
Avg. Step 2 & 94.04\% & \textbf{96.94\%} & 74.67\% & 86.71\% & \textbf{95.81\%} & \textbf{66.21\%} & \textbf{90.18\%} & \textbf{96.69\%} & 75.50\% \\
\bottomrule
\end{tabular}
}
\caption{Scores in consistency within the same definition. In \textbf{bold} the highest values observed (per step), \underline{underlined} the outliers identified with the Interquartile Range method.} 
\label{tab:const_within}

\end{table*}

\section{Sensitivity}\label{app:sensitivity}

In this analysis, instead of comparing the results produced by each run, we are comparing how the answers change definition by definition, in other words, how sensitive is the model to different definitions. 

We represent this trough confusion matrices reflecting the average non-consistent answers between each definition. 
Through this sensitivity analysis we observe that generally all the models tend to be less and less sensitive as more information is added to the definition. 

The same does not apply to the second step, when more specific information are added to the definition (i.e., notion of implicit HS, or possible implications), we instead observe more sensitivity when we are comparing definitions with different Conceptual Elements (i.e., definition with CE of implicit HS vs. definition with CE of Exception), and vice versa when the these CEs are shared by the compared definitions. Especially, this results coherent with what we observe in Sec. \ref{subsec:type_of_hate}, we observe more non-consistent answer when we are comparing definitions with different CEs. For instance, when we are comparing the definition with the CE of exception and the definition with the CE of implicit HS, we observe an higher number different responses, hinting that the model is classifying data-points in a different way, exactly how we saw in our error analysis. Even though in Sec. \ref{subsec:type_of_hate} we could test it only for HateCheck, we observe the same non-consistent pattern in the second step across all three the datasets.

%We calculated the consistency between different %definitions in each of the three runs, Table %\ref{tab:cons_between} reports the averaged values %of the consistency across the three runs. 

%\begin{table*}[h]
%\resizebox{\textwidth}{!}{
%\scriptsize
%\begin{tabular}{@{}l|ccc|ccc|ccc!{}}
%\toprule
%\multicolumn{1}{c|}{\multirow{2}{*}{\textbf{}}} & \multicolumn{3}{c|}{\textbf{HateCheck}} & \multicolumn{3}{c|}{\textbf{LFTW}} & \multicolumn{3}{c}{\textbf{MHS}} \\
%\cmidrule(lr){2-4} \cmidrule(lr){5-7} \cmidrule(lr){8-10}
% & LLama3 & Mistral & FlanT5 & LLama3 & Mistral & FlanT5 & LLama3 & Mistral & FlanT5 \\
%\midrule
%Avg. Step 1  & 83.83\% & 81.97\% & 47.40\% & 70.46\% & 48.47\% & 34.51\% & 76.92\% & 84.84\% & 50.52\% \\
%Std. Step 1 & (±0.05\%) & (±0.08\%) & (±0.71\%) & (±0.13\%) & (±0.43\%) & (±0.53\%) & (±0.20\%) & (±0.01\%) & (±0.65\%) \\
%\midrule
%Avg. Step 2 & 87.75\%	& 83.48\%	& 53.93\%	& 75.00\%	& 82.79\%	& 42.85\%	& 81.90\%	& 87.93\%	& 55.74\% \\
%Std. Step 2 & (±0.11\%) & (±0.17\%) & (±0.74\%) & (±0.38\%) & (±0.12\%) & (±0.30\%) & (±0.42\%) & (±0.01\%) & (±0.27\%) \\
%\bottomrule
%\end{tabular}}
%\caption{Average Consistency and Standard Deviation across all the condition (definition prompted) of each of the three runs.}
%\label{tab:cons_between}
%\end{table*}

\begin{figure}[h]
    \centering
    \begin{subfigure}{0.45\textwidth}
        \includegraphics[width=\linewidth]{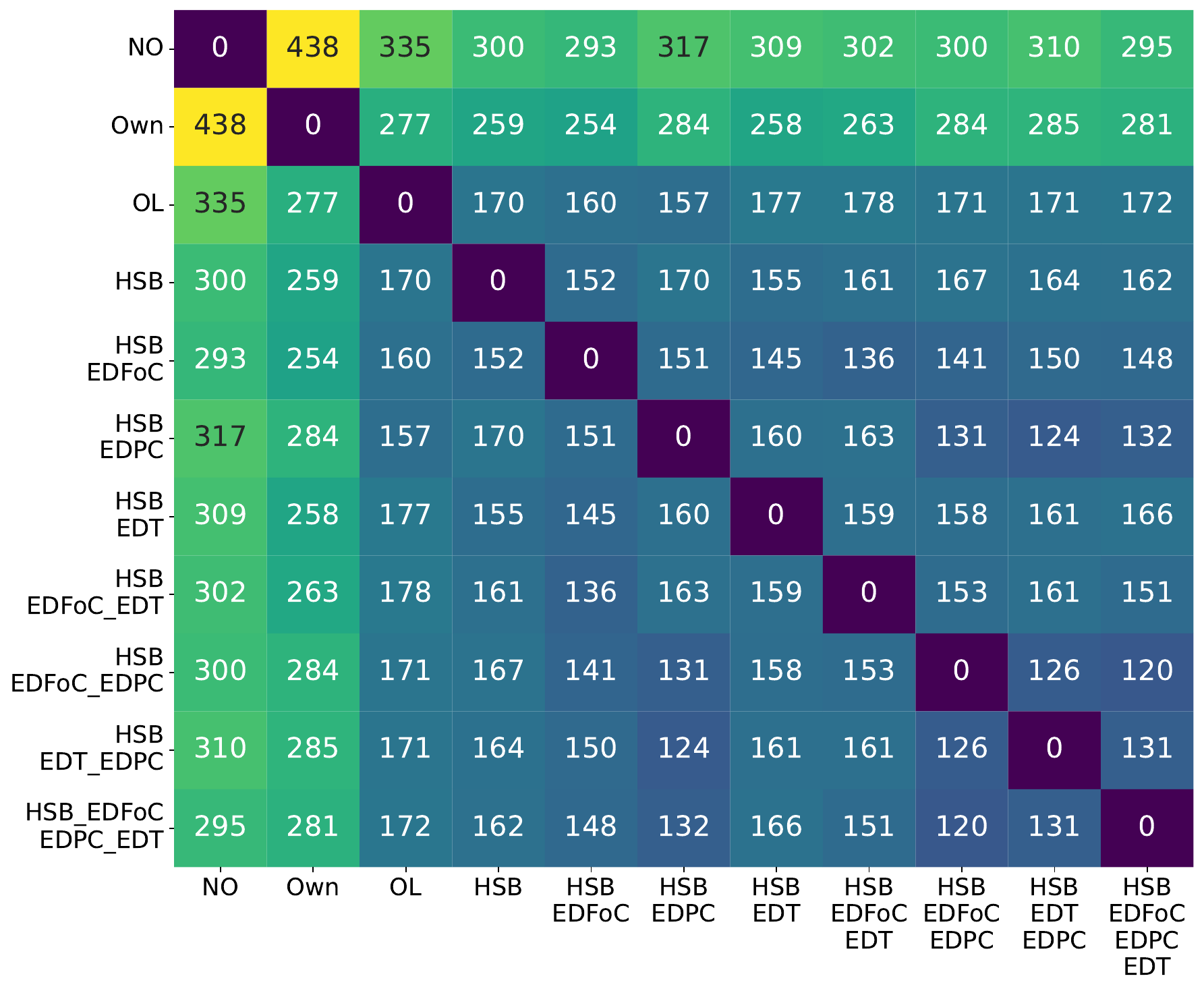}
        \caption{HateCheck Dataset Step 1}
    \end{subfigure}
    \hfill
    \begin{subfigure}{0.45\textwidth}
        \includegraphics[width=\linewidth]{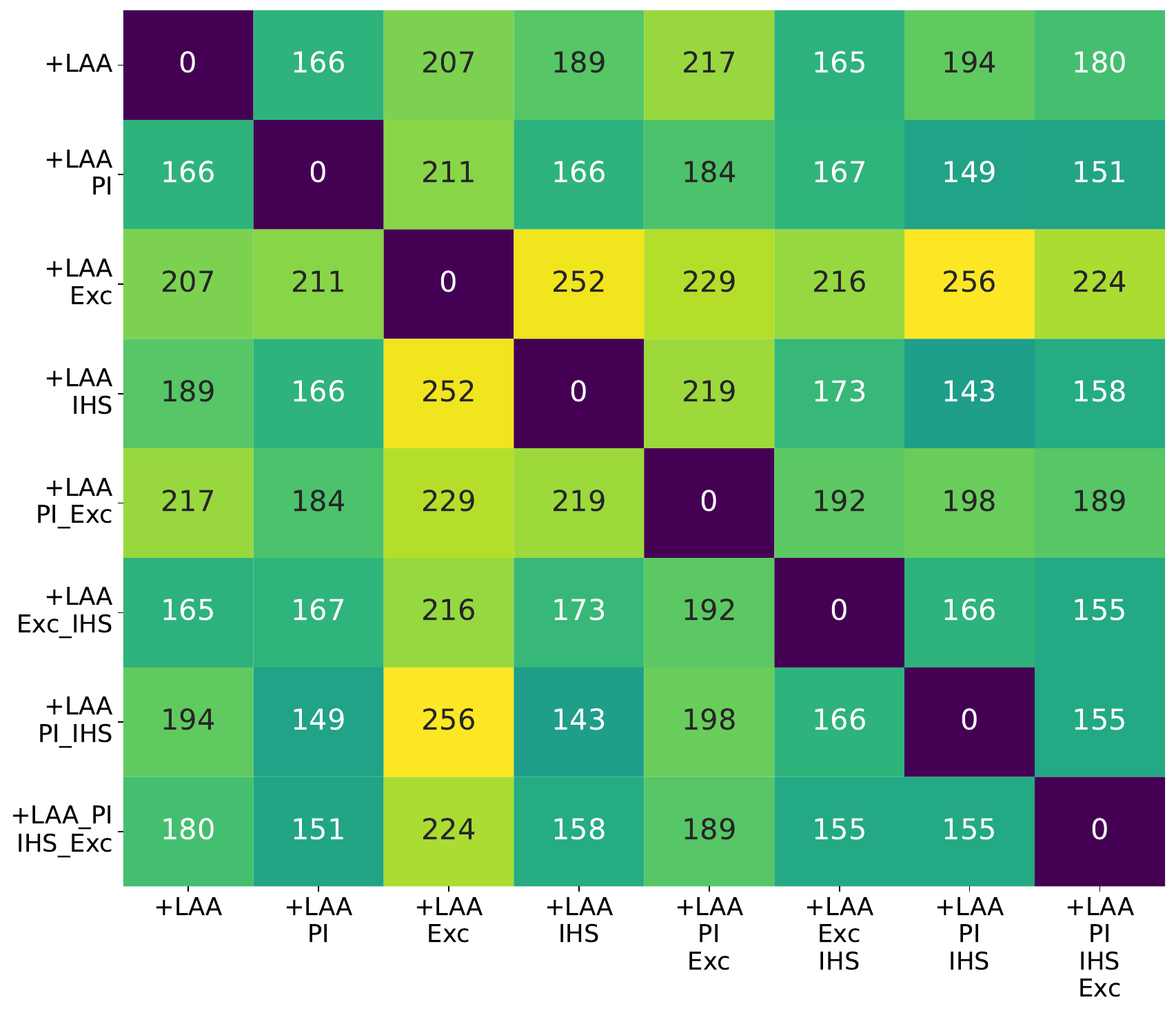}
        \caption{HateCheck Dataset Step 2}
    \end{subfigure}
    
    \vspace{1em} % Add vertical space between rows
    
    \begin{subfigure}{0.45\textwidth}
        \includegraphics[width=\linewidth]{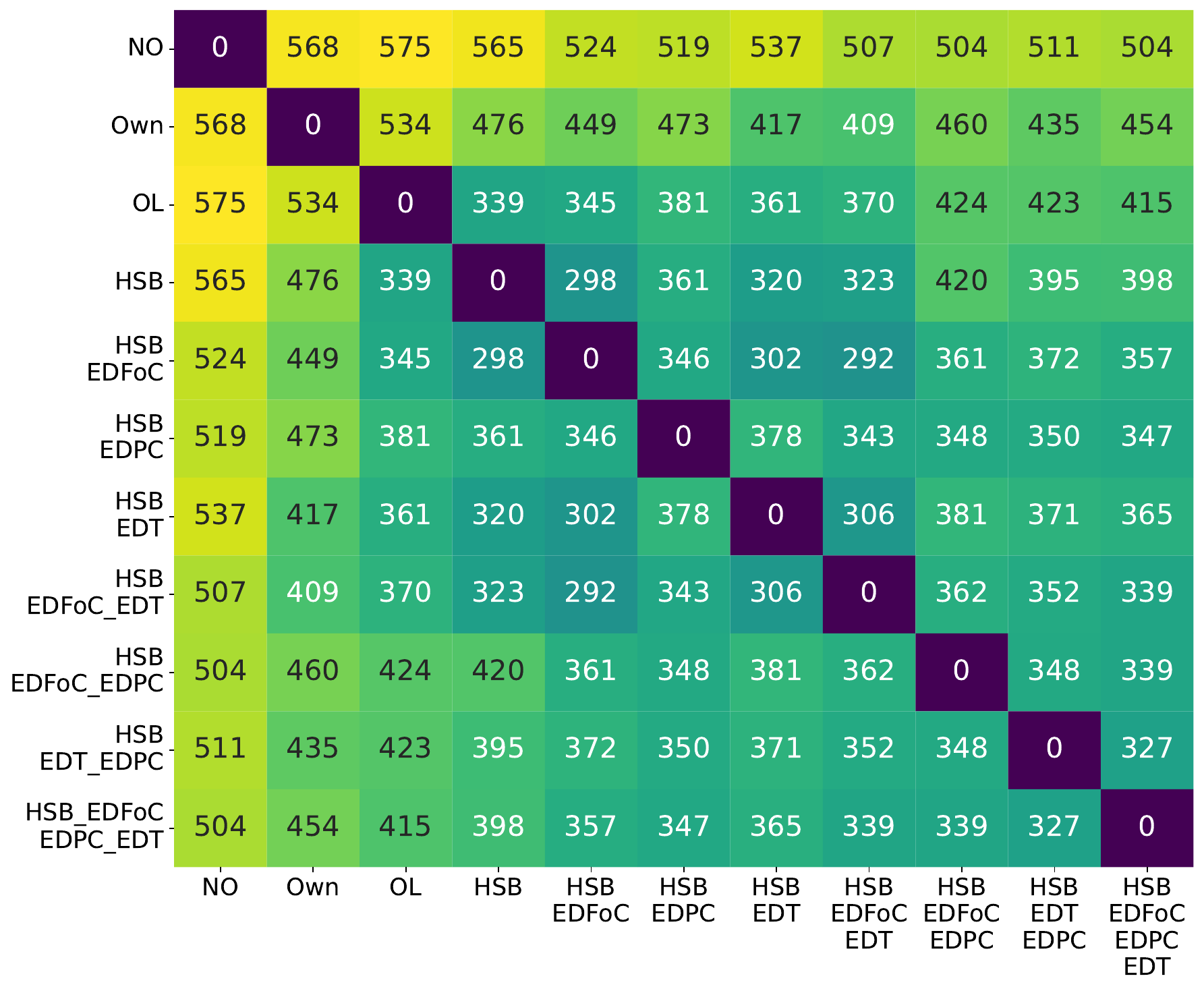}
        \caption{Learning from the Worst Dataset Step 1}
    \end{subfigure}
    \hfill
    \begin{subfigure}{0.45\textwidth}
        \includegraphics[width=\linewidth]{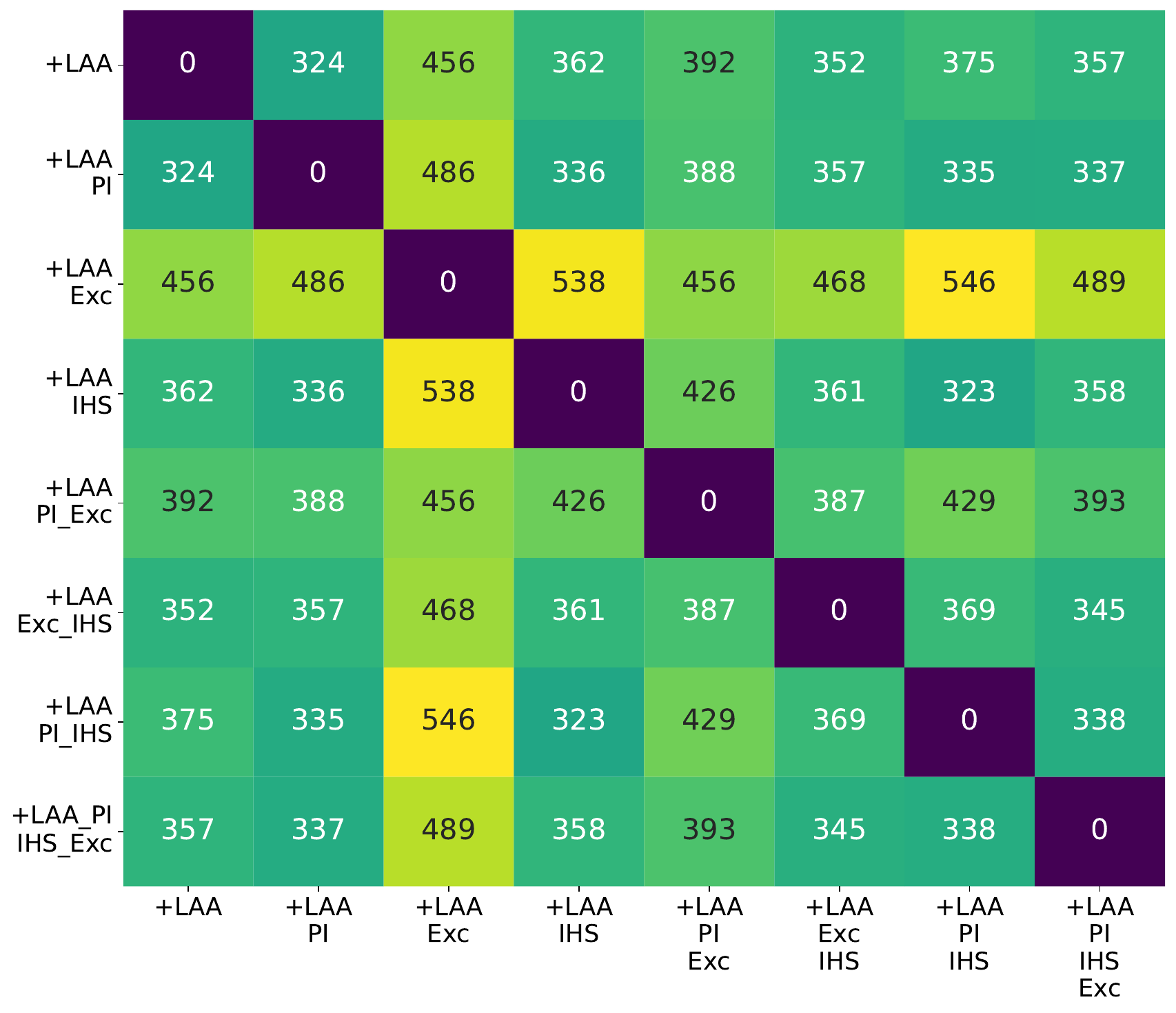}
        \caption{Learning from the Worst Dataset Step 2}
    \end{subfigure}

    \vspace{1em} % Add vertical space between rows

        \centering
    \begin{subfigure}{0.45\textwidth}
        \includegraphics[width=\linewidth]{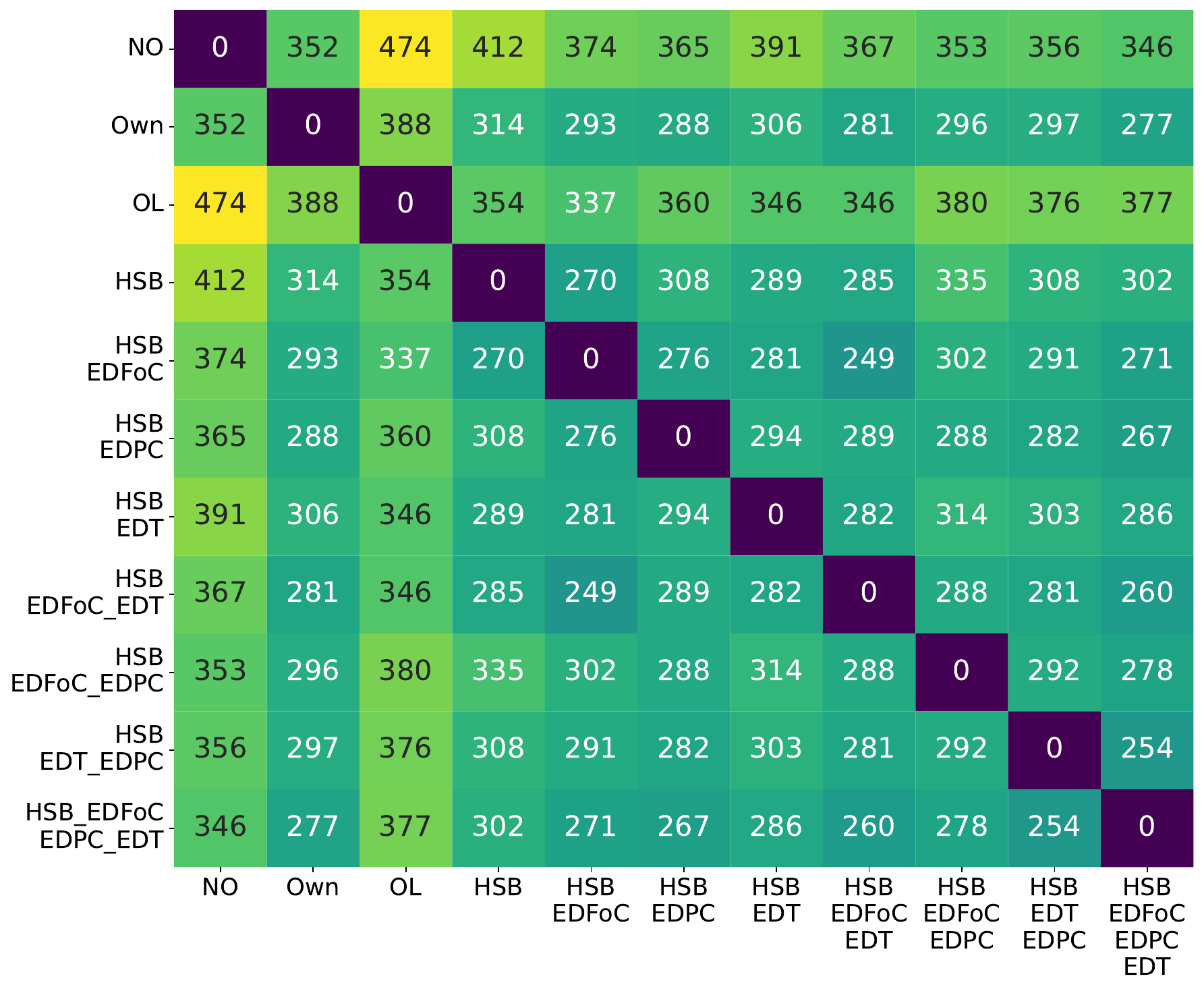}
        \caption{Measuring Hate Speech Dataset Step 1}
    \end{subfigure}
    \hfill
    \begin{subfigure}{0.45\textwidth}
        \includegraphics[width=\linewidth]{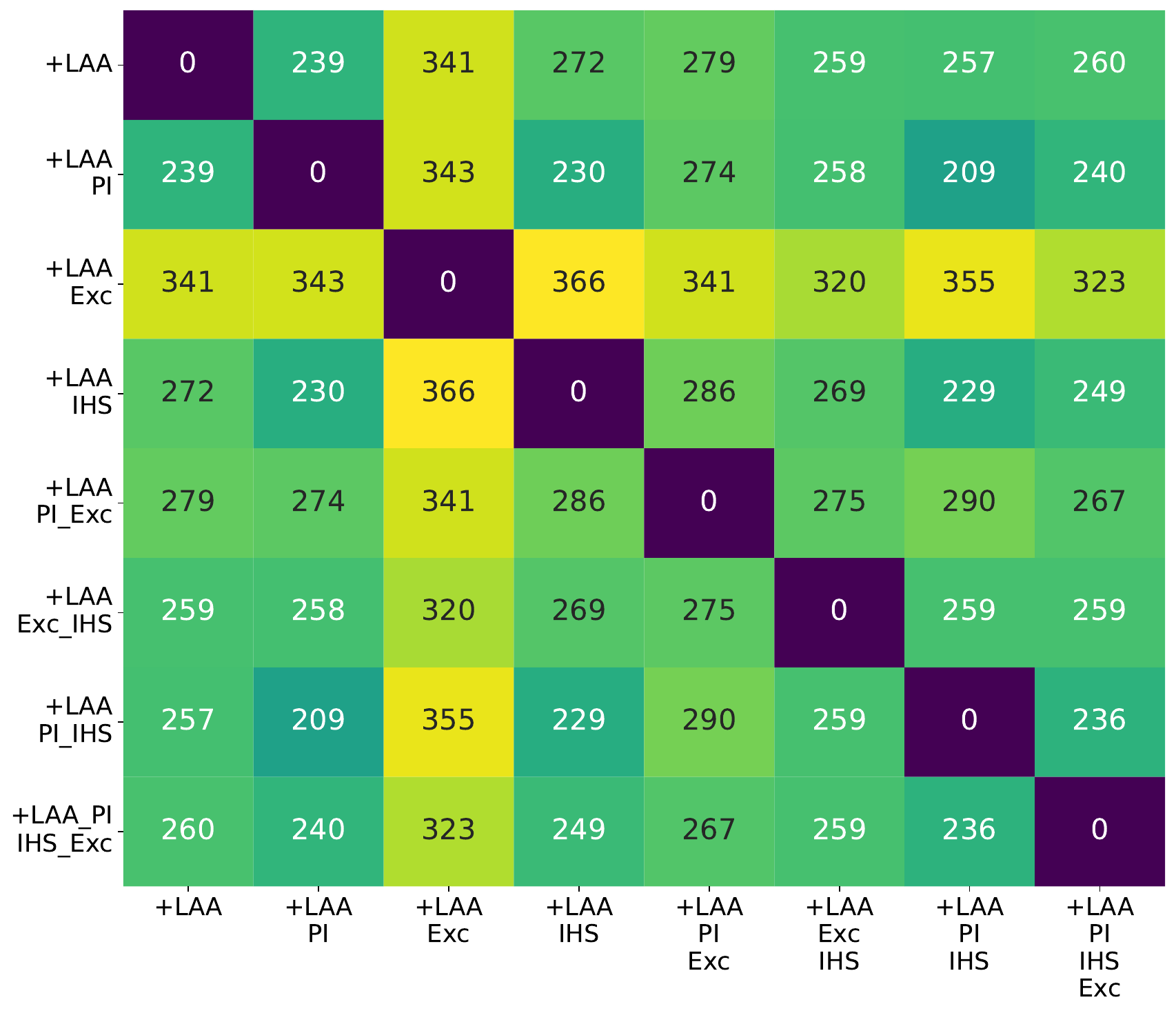}
        \caption{Measuring Hate Speech Dataset Step 2}
    \end{subfigure}
    \caption{Confusion matrices of non-consistent answer between definitions in \textbf{LLama3}}
    \label{fig:allfiguresLLama}
\end{figure}

\begin{figure}[htbp]
    \centering
 \begin{subfigure}{0.45\textwidth}
        \includegraphics[width=\linewidth]{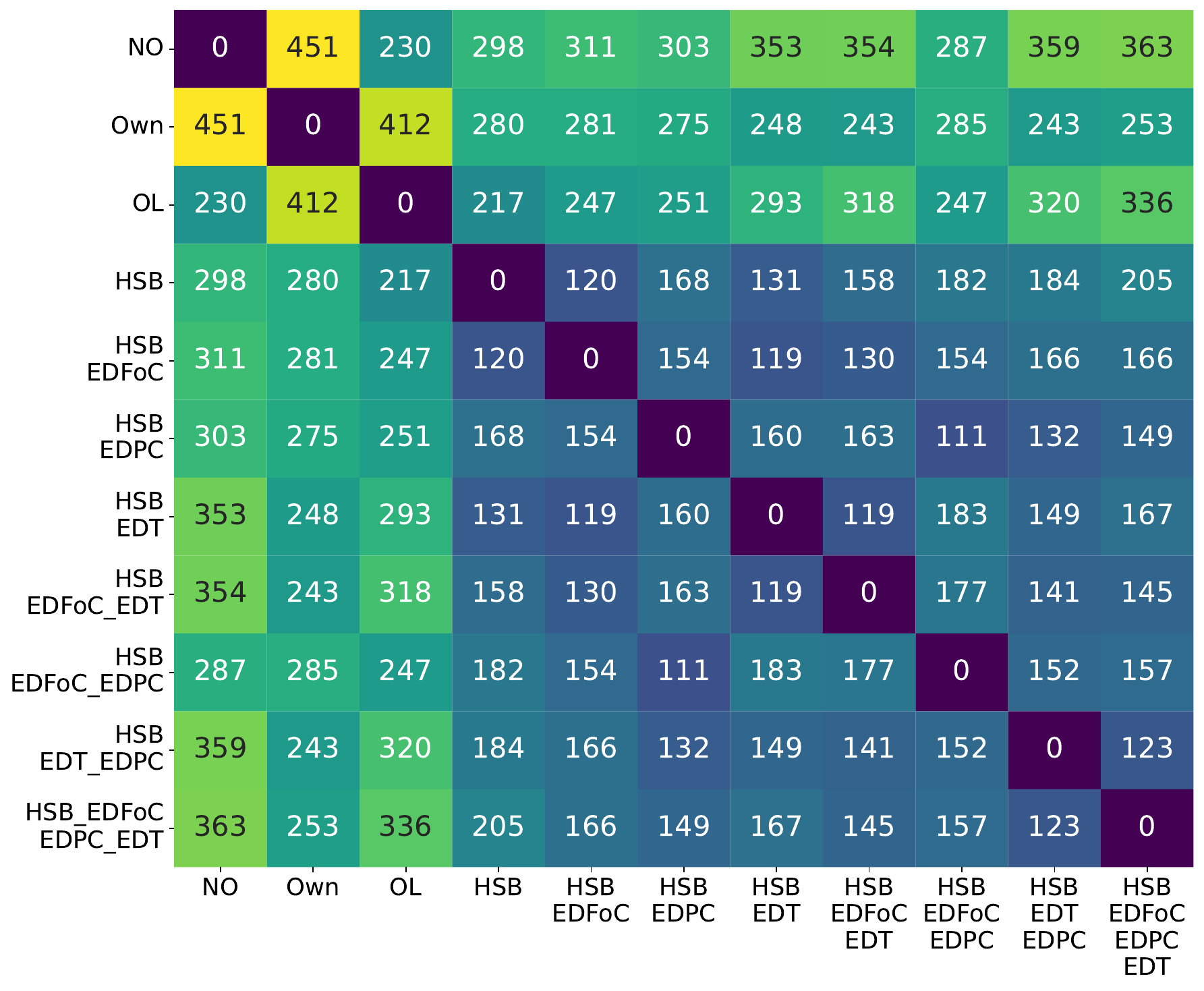}
        \caption{HateCheck Dataset Step 1}
    \end{subfigure}
    \hfill
    \begin{subfigure}{0.45\textwidth}
        \includegraphics[width=\linewidth]{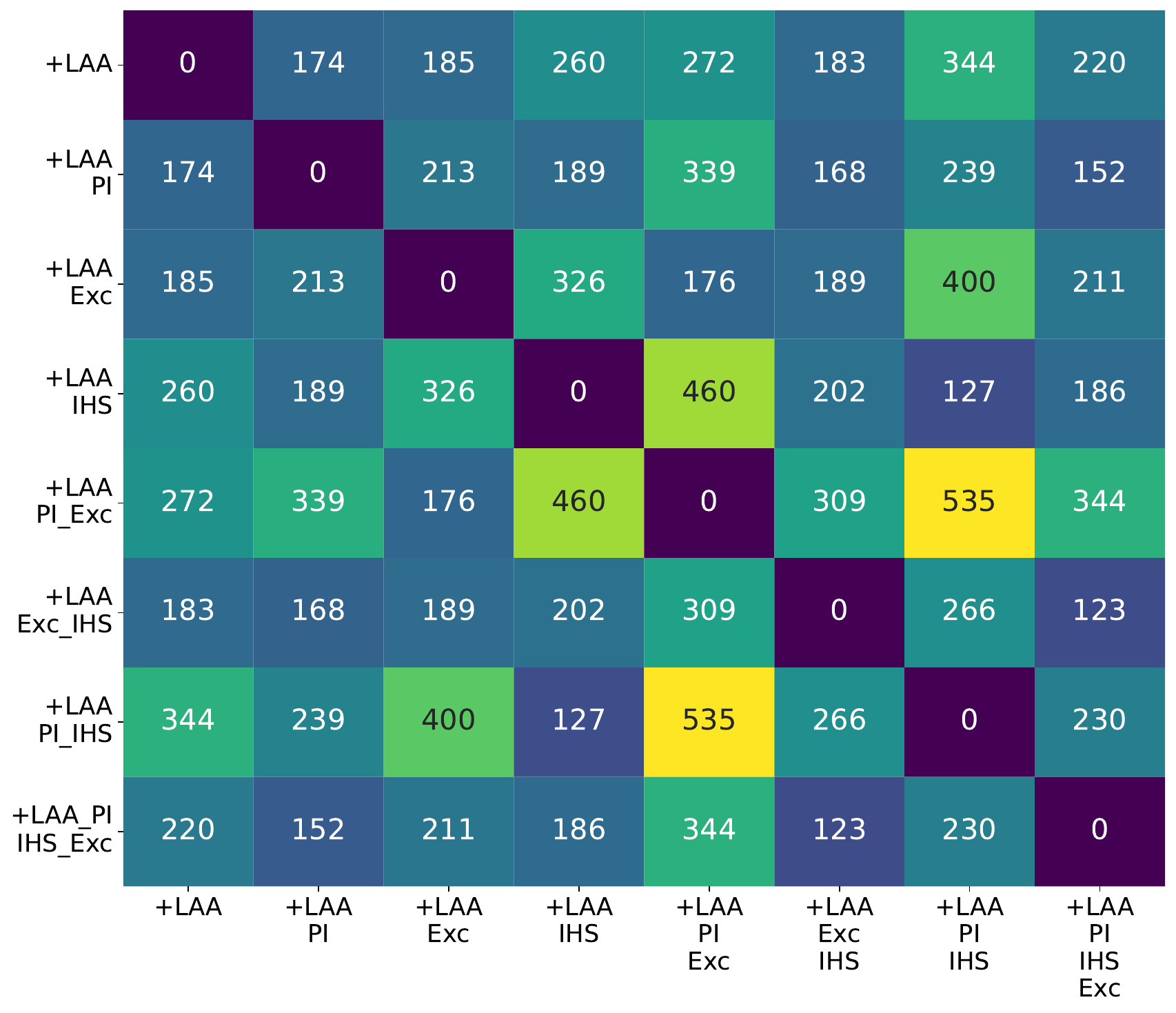}
        \caption{HateCheck Dataset Step 2}
    \end{subfigure}

    \vspace{1em} % Add vertical space between rows

        \centering
    \begin{subfigure}{0.45\textwidth}
        \includegraphics[width=\linewidth]{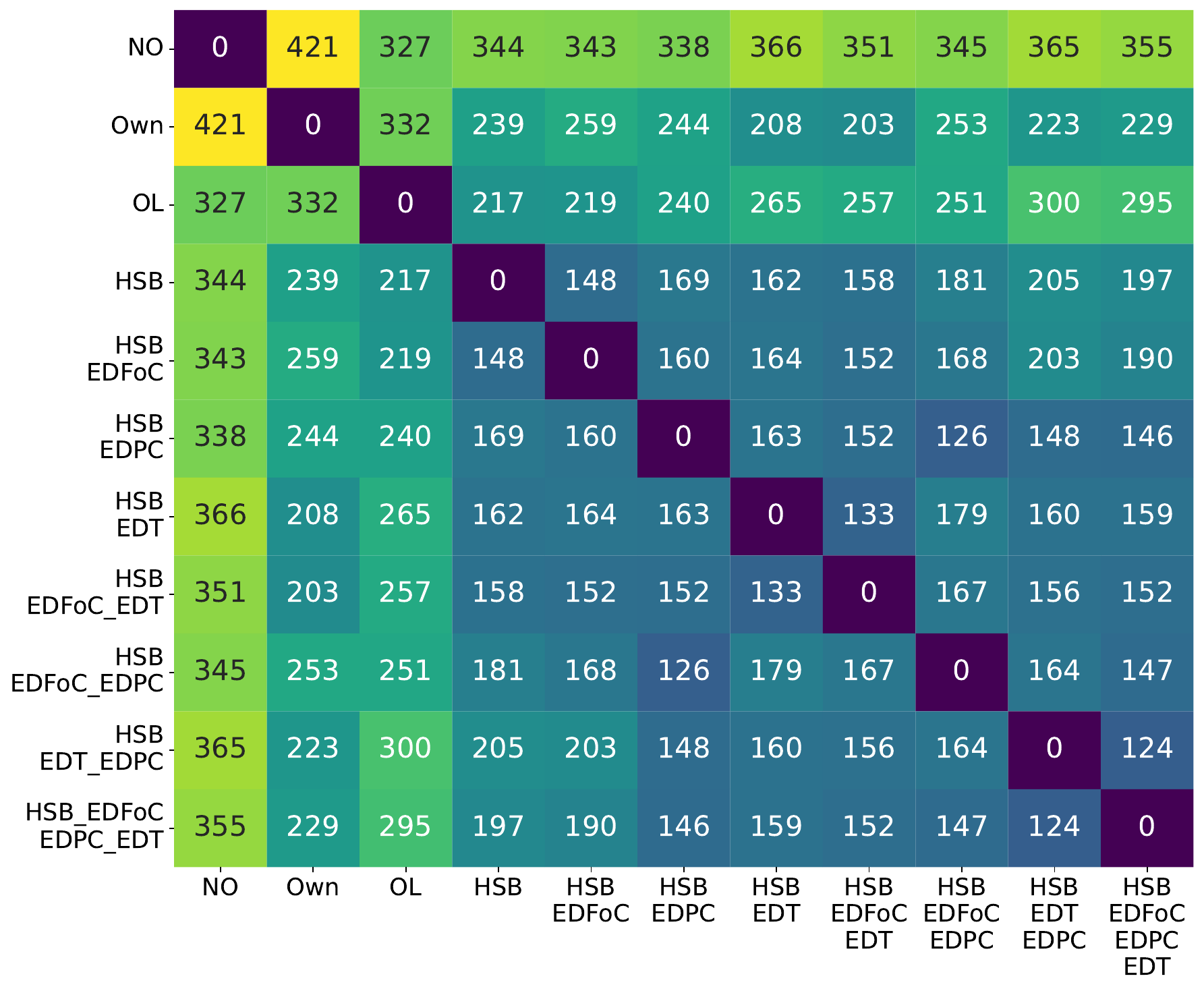}
        \caption{Learning from the Worst Dataset Step 1}
    \end{subfigure}
    \hfill
    \begin{subfigure}{0.45\textwidth}
        \includegraphics[width=\linewidth]{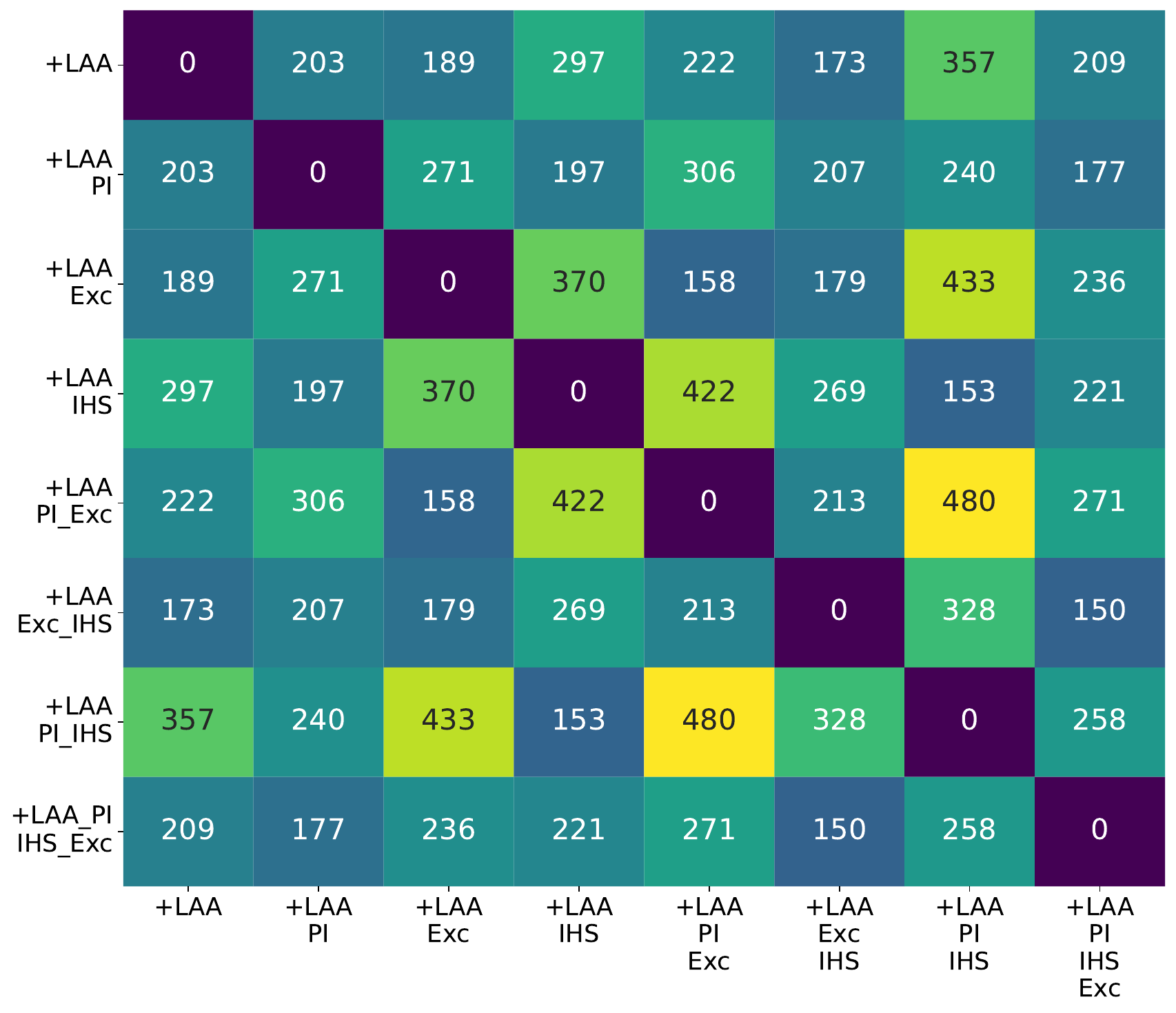}
        \caption{Learning from the Worst Dataset Step 2}
    \end{subfigure}

    \vspace{1em} % Add vertical space between rows

        \begin{subfigure}{0.45\textwidth}
        \includegraphics[width=\linewidth]{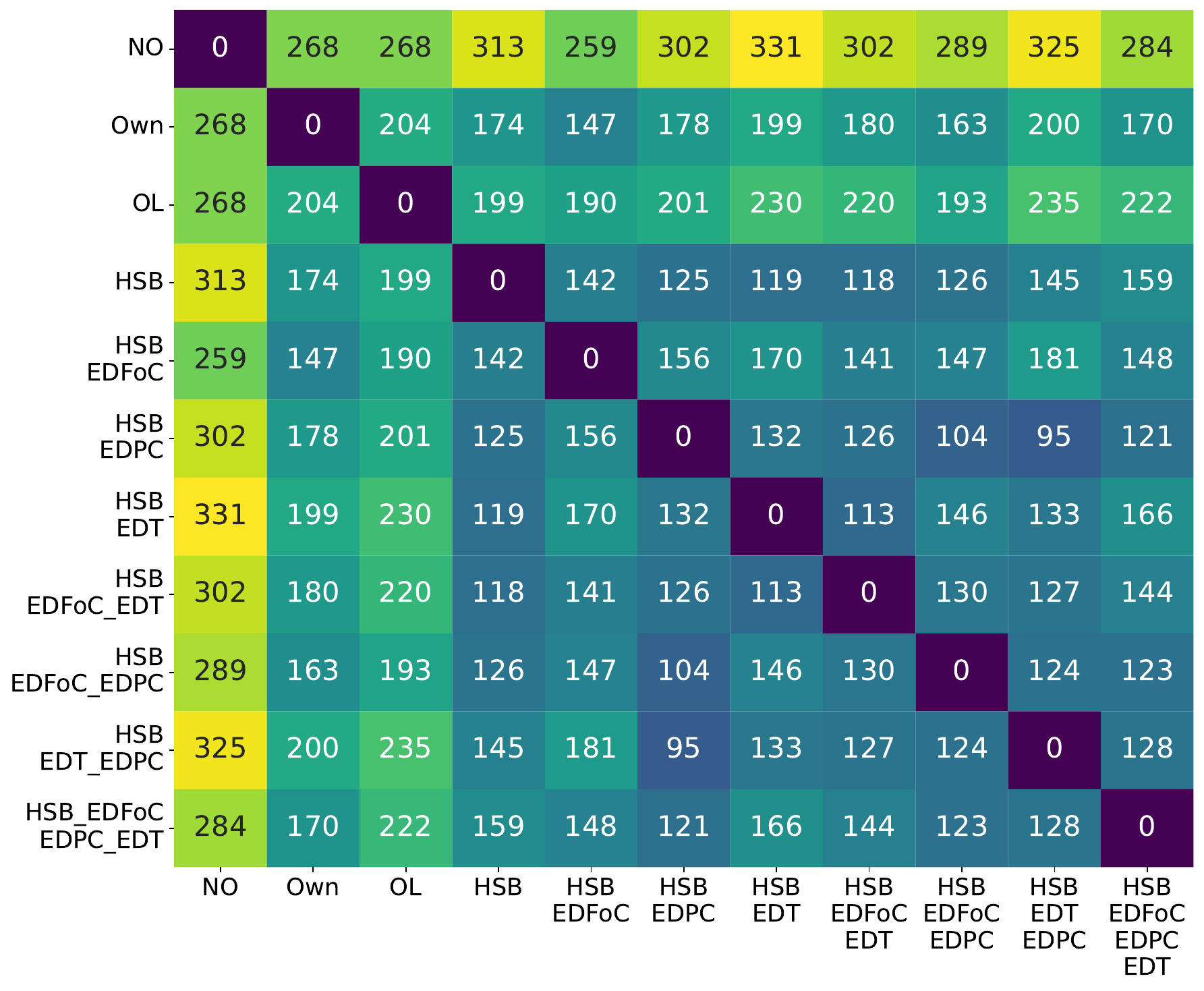}
        \caption{Measuring Hate Speech Dataset Step 1}
    \end{subfigure}
    \hfill
    \begin{subfigure}{0.45\textwidth}
        \includegraphics[width=\linewidth]{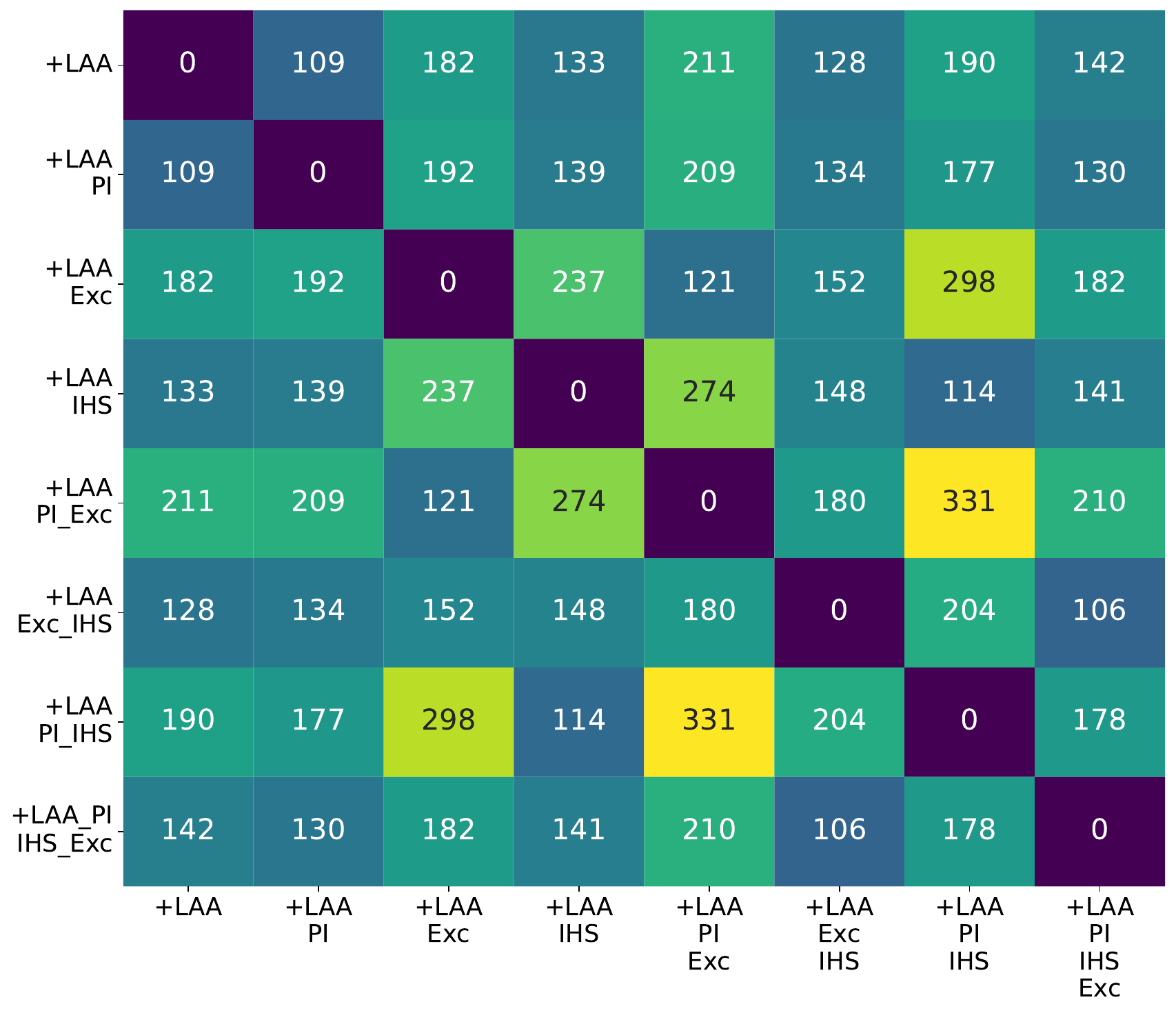}
        \caption{Measuring Hate Speech Dataset Step 2}
    \end{subfigure}

    \caption{Confusion matrices of non-consistent answer between definitions in \textbf{Mistral}}
    \label{fig:allfiguresMistral}
\end{figure}

\begin{figure}[htbp]
    \centering
 \begin{subfigure}{0.45\textwidth}
        \includegraphics[width=\linewidth]{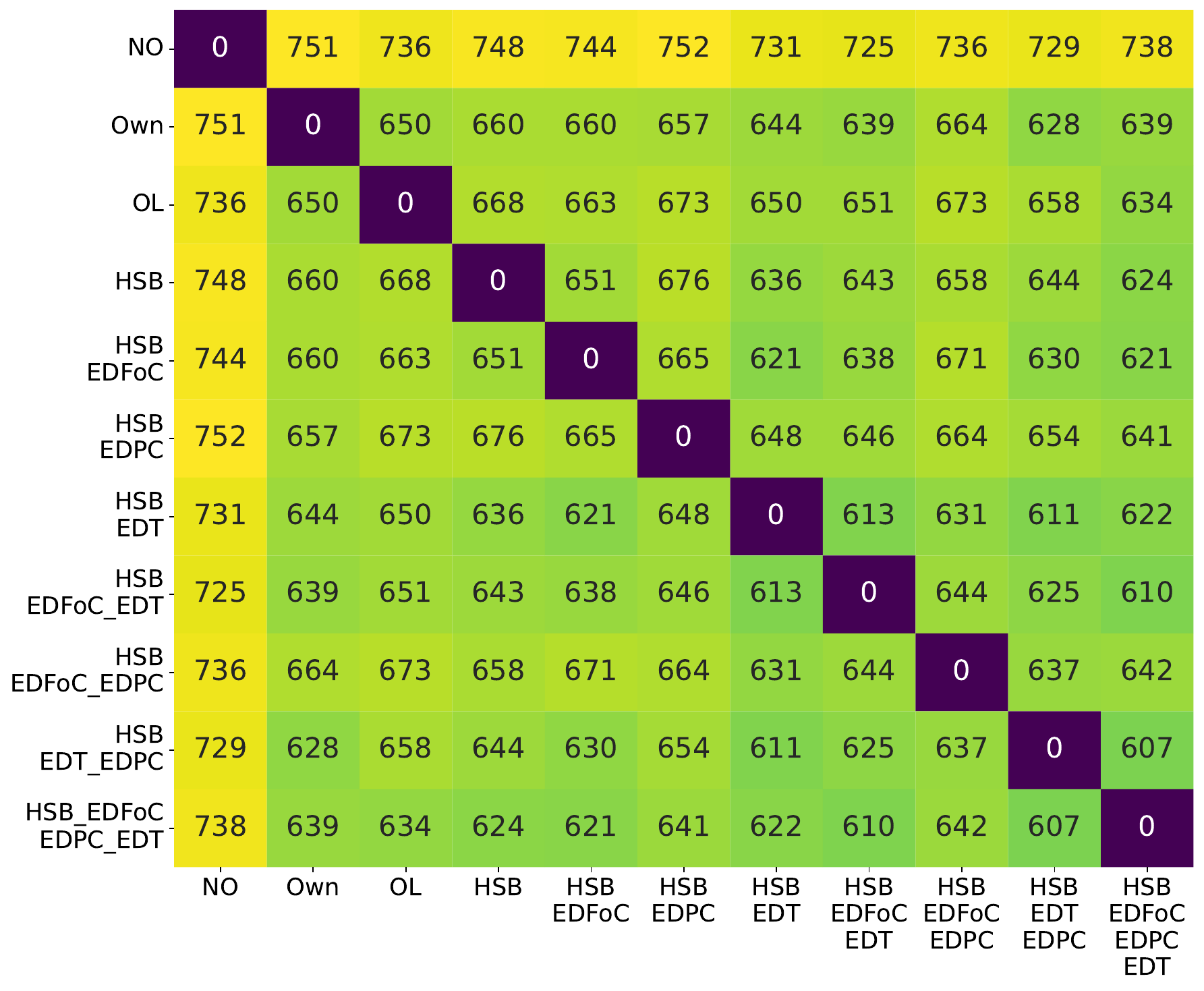}
        \caption{HateCheck Dataset Step 1}
    \end{subfigure}
    \hfill
    \begin{subfigure}{0.45\textwidth}
        \includegraphics[width=\linewidth]{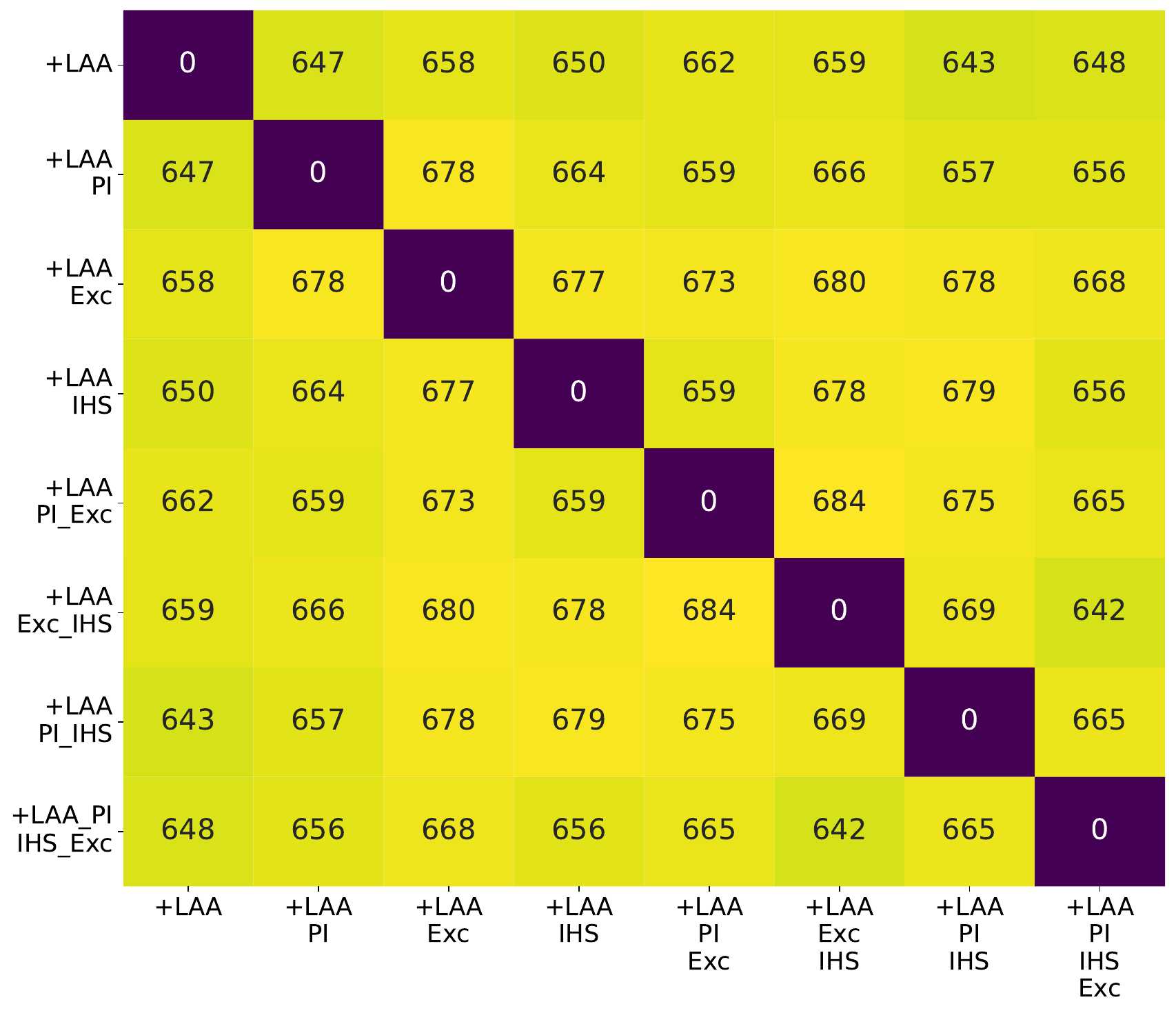}
        \caption{HateCheck Dataset Step 2}
    \end{subfigure}

    \vspace{1em} % Add vertical space between rows

        \centering
    \begin{subfigure}{0.45\textwidth}
        \includegraphics[width=\linewidth]{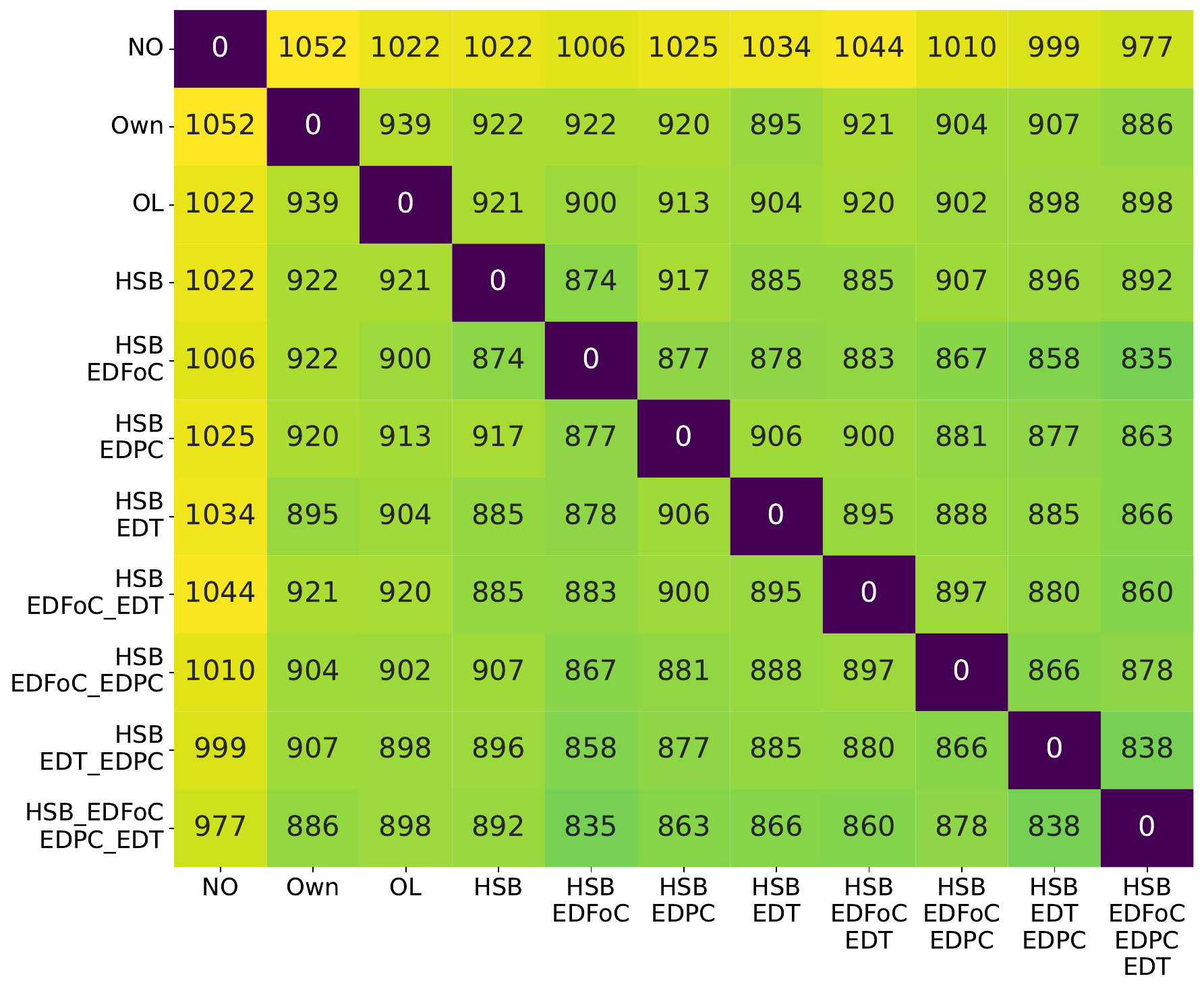}
        \caption{Learning from the Worst Dataset Step 1}
    \end{subfigure}
    \hfill
    \begin{subfigure}{0.45\textwidth}
        \includegraphics[width=\linewidth]{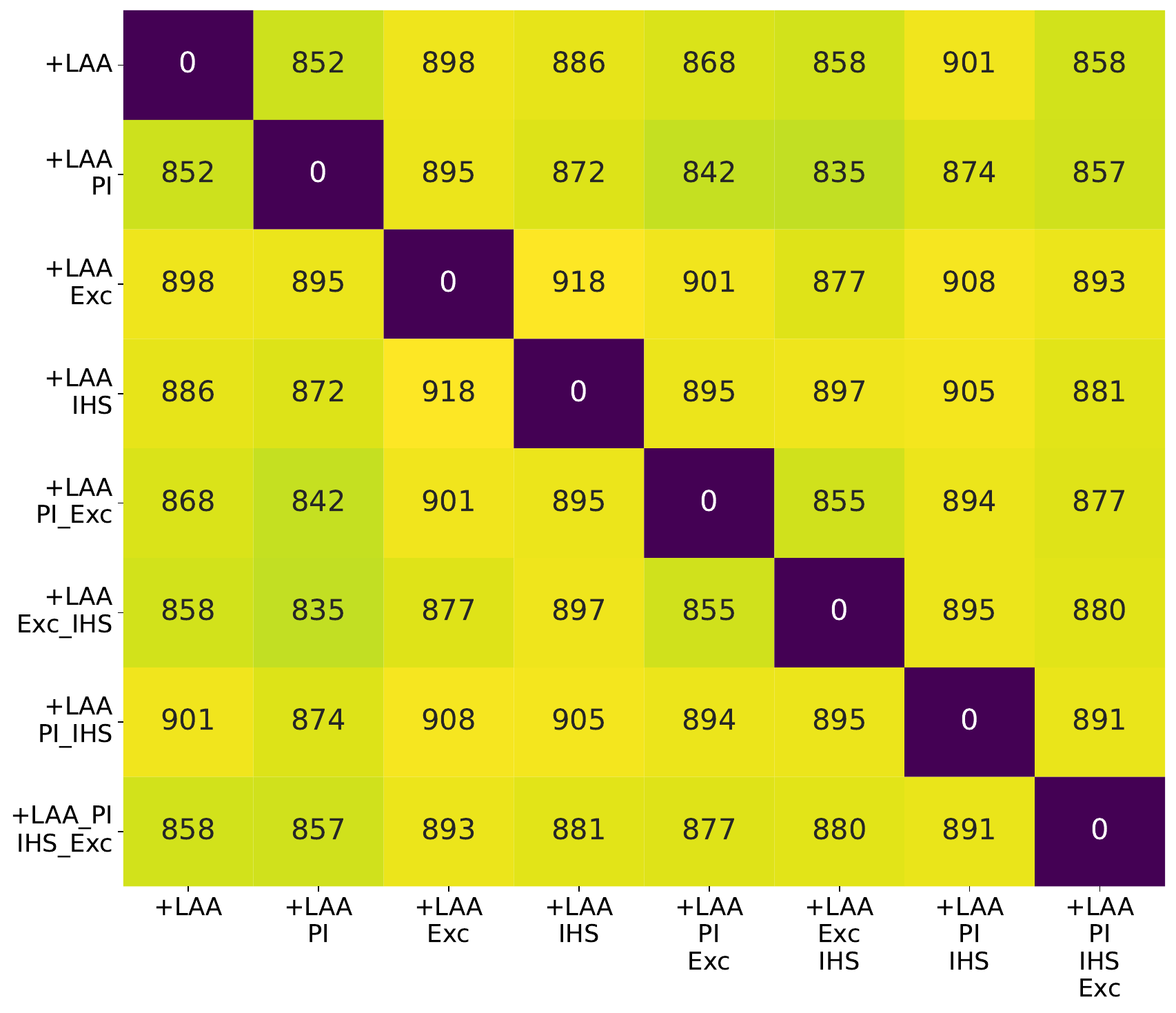}
        \caption{Learning from the Worst Dataset Step 2}
    \end{subfigure}

    \vspace{1em} % Add vertical space between rows

        \begin{subfigure}{0.45\textwidth}
        \includegraphics[width=\linewidth]{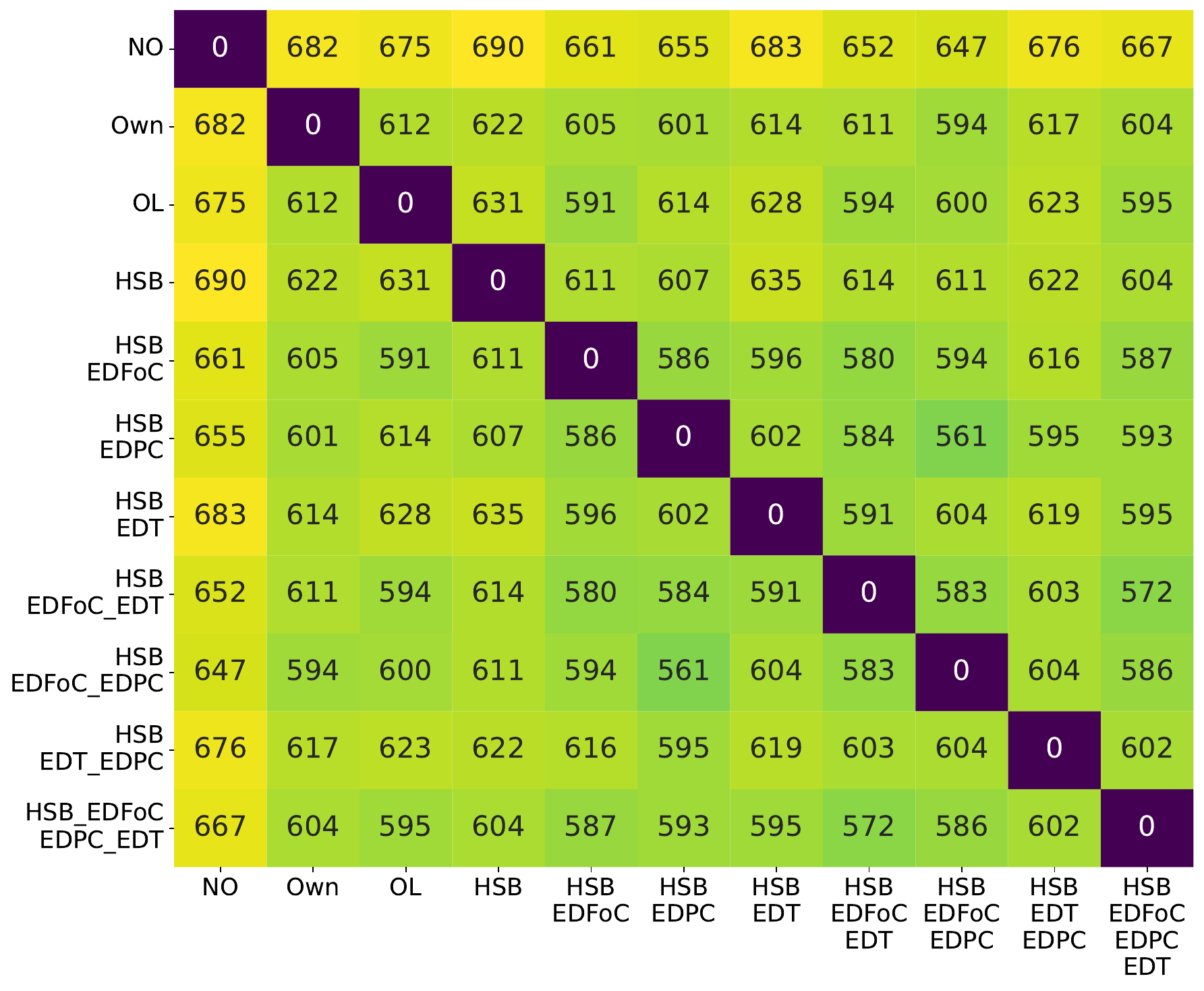}
        \caption{Measuring Hate Speech Dataset Step 1}
    \end{subfigure}
    \hfill
    \begin{subfigure}{0.45\textwidth}
        \includegraphics[width=\linewidth]{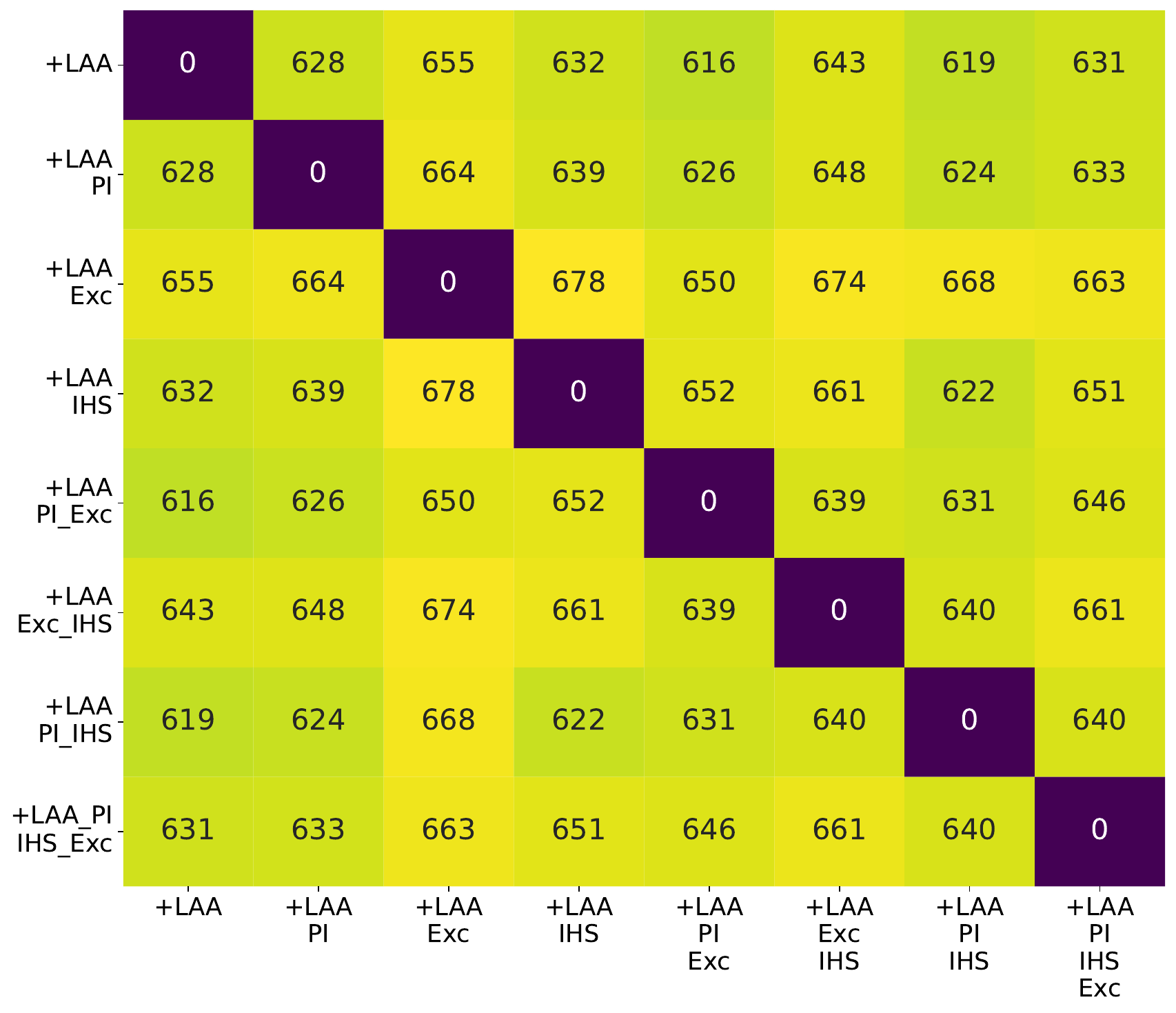}
        \caption{Measuring Hate Speech Dataset Step 2}
    \end{subfigure}

    \caption{Confusion matrices of non-consistent answer between definitions in \textbf{Flan-T5}}
    \label{fig:allfiguresFlan}
\end{figure}

\clearpage
\section{Error Distribution Based on Classes}\label{app:Error_anal_classes}

\begin{figure*}[htbp]
    \centering
    \includegraphics[width=\textwidth]{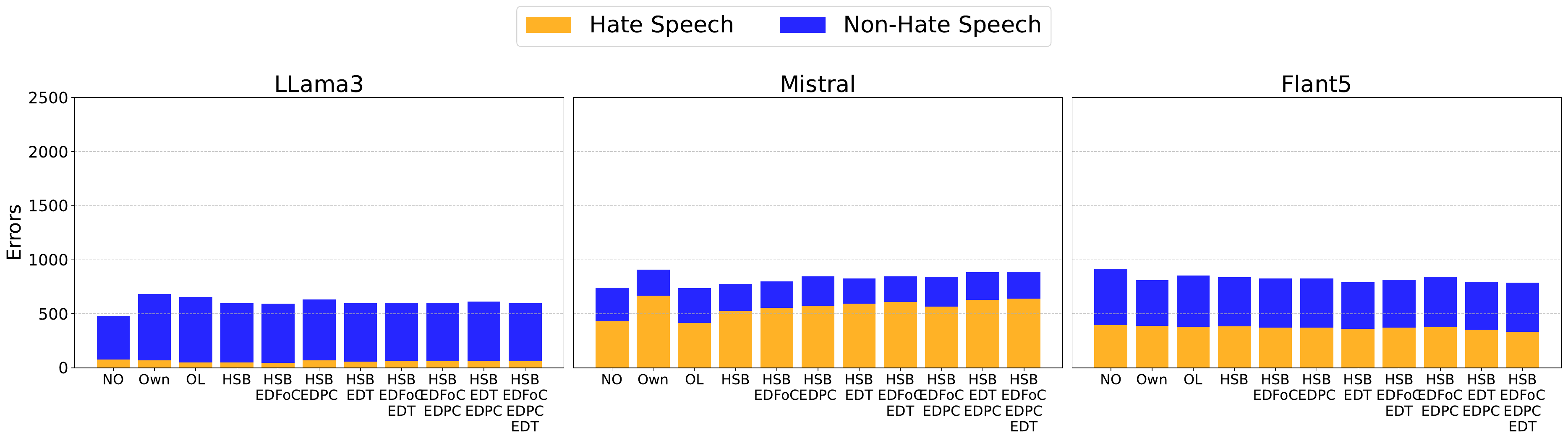}
    \caption{Distribution of errors across the three models on HateCheck (Step 1).}
\end{figure*}

\begin{figure*}[htbp]
    \centering
    \includegraphics[width=\textwidth]{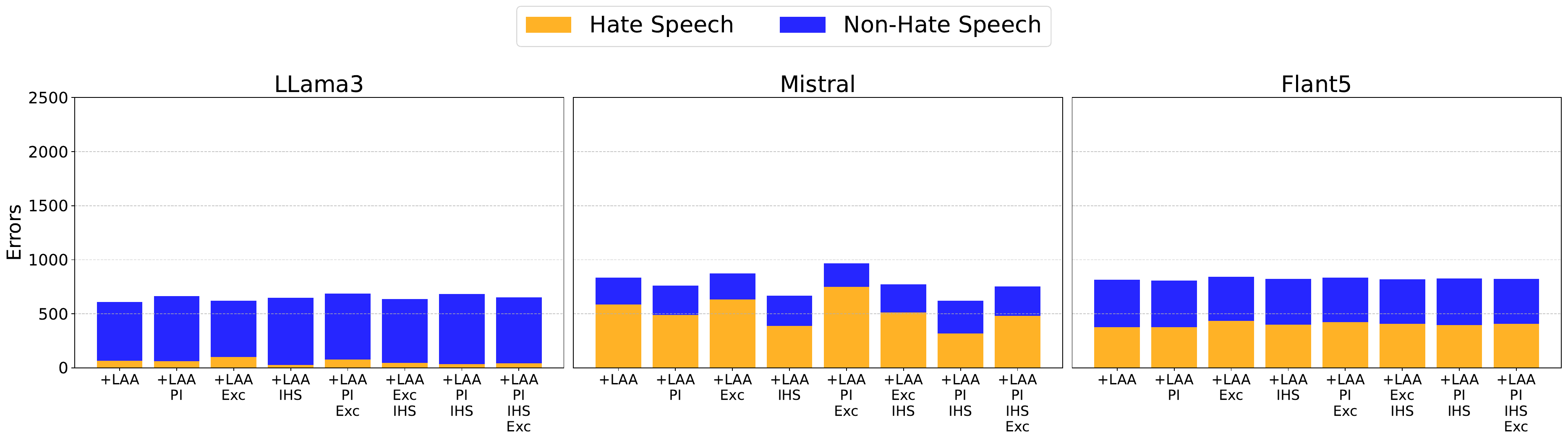}
    \caption{Distribution of errors across the three models on HateCheck (Step 2).}
\end{figure*}

\begin{figure*}[htbp]
    \centering
    \includegraphics[width=\textwidth]{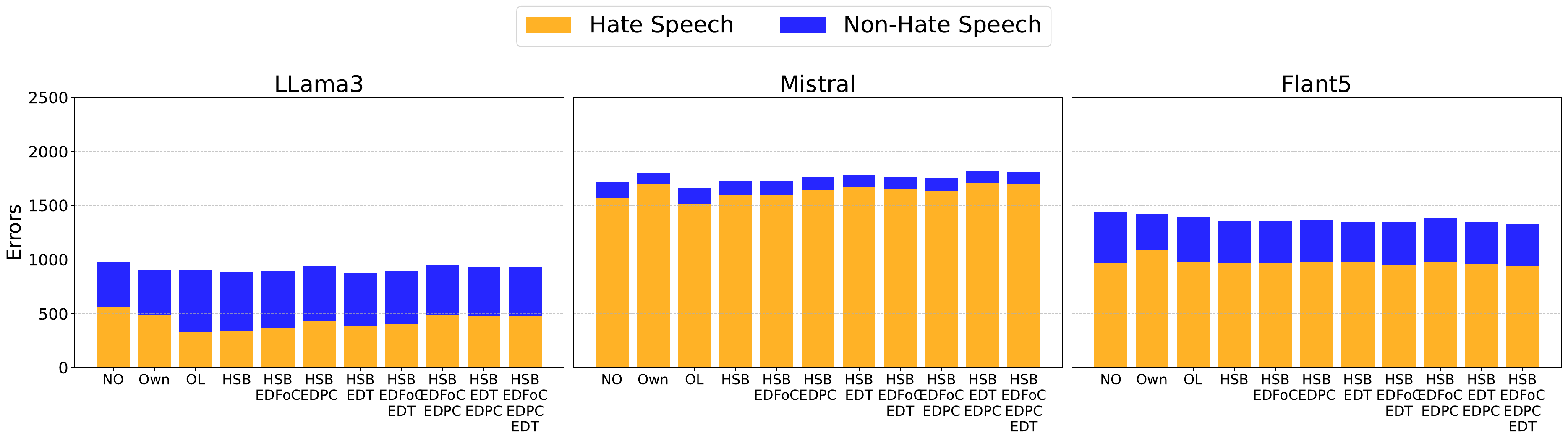}
    \caption{Distribution of errors across the three models on Learning from the Worst (Step 1).}
\end{figure*}

\begin{figure*}[htbp]
    \centering
    \includegraphics[width=\textwidth]{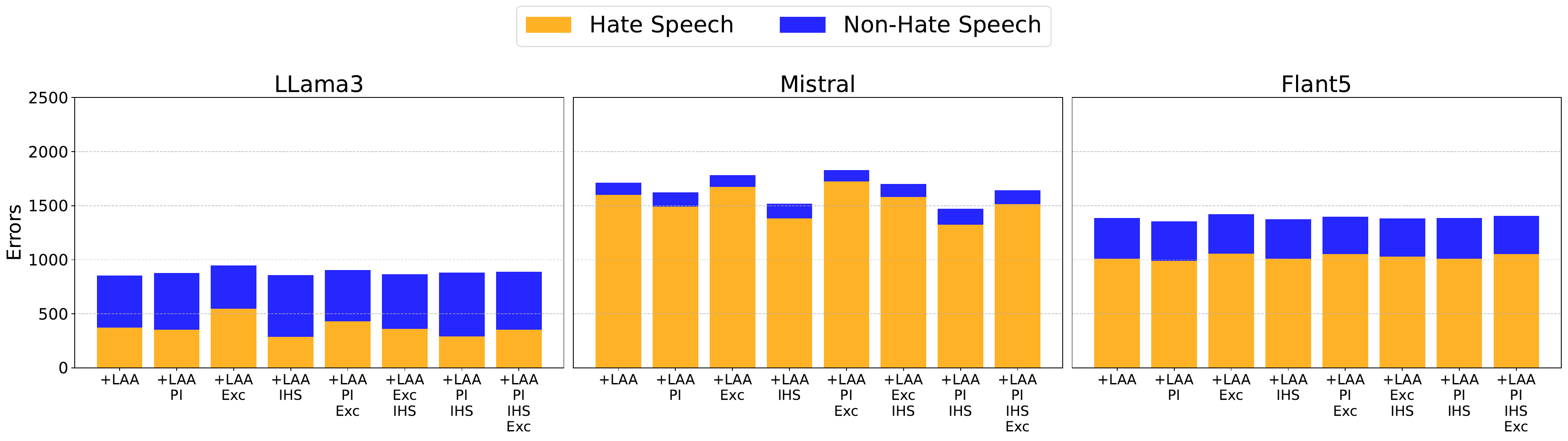}
    \caption{Distribution of errors across the three models on Learning from the Worst (Step 2).}
\end{figure*}

\begin{figure*}[htbp]
    \centering
    \includegraphics[width=\textwidth]{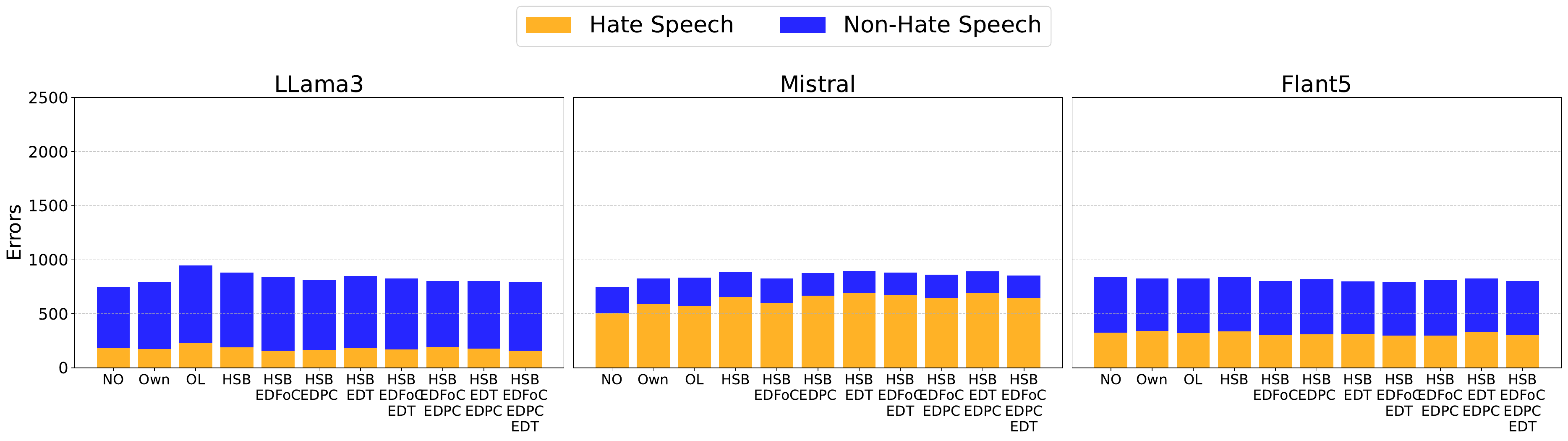}
    \caption{Distribution of errors across the three models on Measuring Hate Speech (Step 1).}
\end{figure*}

\begin{figure*}[htbp]
    \centering
    \includegraphics[width=\textwidth]{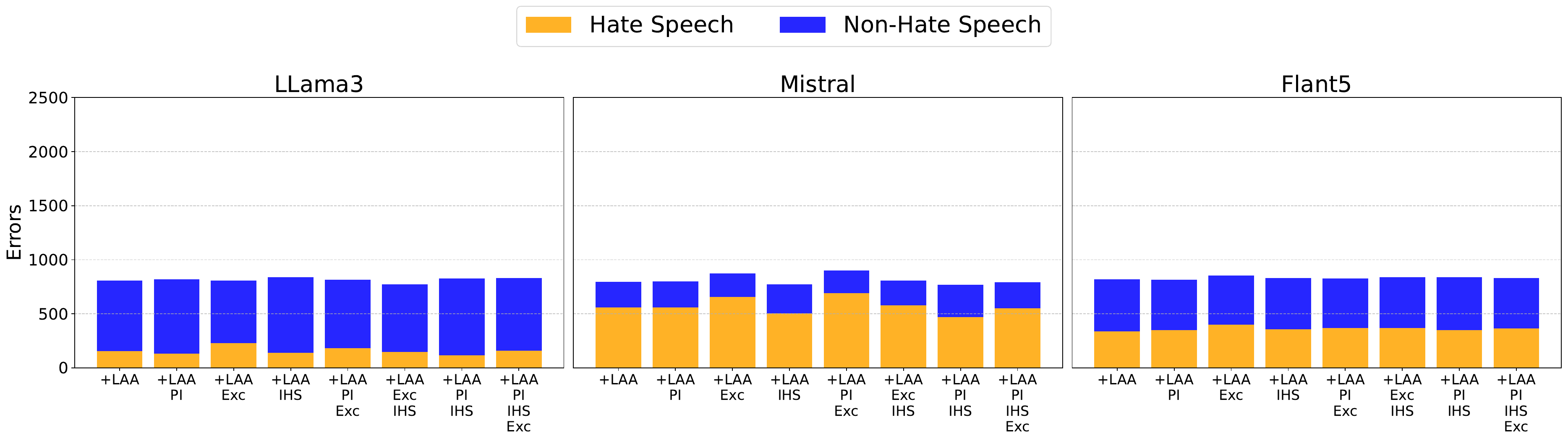}
    \caption{Distribution of errors across the three models on Measuring Hate Speech (Step 2).}
\end{figure*}

\clearpage
\section{Grouping functionalities in Macro Classes}\label{app:functionalities}

HateCheck dataset presents 29 functionalities (or type of hate) "motivated by a review of previous research and a series of interviews with civil society stakeholders" \cite{rottger2021hatecheck}. In order to investigate trends after the inference phase, we decided to cluster these into 5 Macro Classes reflecting:

\begin{itemize}
    \item \underline{Hate Speech:} data-points that simply reflect direct and evident Hate Speech, such as: (slur, dehumanization, threats, etc...)
    \item \underline{Non Hate Speech:} data-points that reflect general sentences without conveying hate (i.e. "I am a trans person", "No fucking way")
    \item \underline{Leet Hate Speech:} data-points build to resemble \textit{leetspeak}, so containing different spelling errors, which makes them more difficult to be detected, for instance: "Ihate trans people" or "I h4te immigrants".
    \item \underline{Misleading Not Hate Speech:} data-points which do not convey HS but are built in a way which resemble an Hate Speech comment such as: "Saying that you hate women shows how misguided you are" or "She ought to be executed".
    \item \underline{Special Hate Speech:} data-points which are more implicitly phrased such as: "Never trust a single word said by a woman" or "I really do believe black people are just lesser beings".
\end{itemize}

\begin{table}[h]
    \centering
    \begin{tabular}{lp{0.23\columnwidth}}
        \toprule
        Macro-classes       & Functionalities \\
        \midrule
        \highlightred{Hate Speech (HS)} &  derog\_dehum\_h \newline derog\_neg\_attrib\_h \newline threat\_dir\_h \newline negate\_pos\_h \newline derog\_neg\_emote\_h \newline slur\_h \newline ref\_subs\_sent\_h \newline threat\_norm\_h \newline  profanity\_h \\
        \midrule
        \highlightgreen{Non Hate Speech (NHS)} & ident\_pos\_nh \newline ident\_neutral\_nh \newline profanity\_nh \newline target\_obj\_nh \\
        \midrule
        \highlightorange{Leet HS}  & space\_deleted, \newline leet\_hate\_speech, \newline character\_swap, \newline space\_add, \newline character\_deleted \\
        \midrule
        \highlightblue{Misleading NHS}  & counter\_ref\_nh \newline negate\_neg\_nh \newline counter\_quote\_nh \newline target\_indiv\_nh \newline target\_group\_nh \newline slur\_reclaimed\_nh \newline slur\_homonym\_nh \\
        \midrule
        \highlightviolet{Special HS} & derog\_impl\_h \newline ref\_subs\_clause\_h \newline phrase\_question\_h \newline phrase\_opinion\_h \\
        \bottomrule
    \end{tabular}
    \caption{29 functionalities (type of hate) grouped in 5 Macro Classes.}
\end{table}

\newpage
\section{HateCheck Errors - Functionalities}\label{app:fun_gen}

%Here below in Table \ref{tab:fun_gen} we report the average error per type of hate for each model, this provides an overview of each model's behaviour. We first calculated the error ratio for each functionality by summing the errors for each definition of the first Step (where there are not CE which can possibly influence the inference step), for then weighting them according to the number of instances per type of hate in HateCheck, which ranges from 30 to 210.

Below, we report the relative average model errors across all HateCheck functionalities.
\\

\noindent
%We then observe more in details the definition effect on each macro class. In Table \ref{tab:funB&W} we report the best and worst definition in terms of errors for each model.

\begin{table}[H]
\centering
\begin{tabular}{cc|cc|cc}
\toprule
\multicolumn{2}{c|}{LLama-3} & \multicolumn{2}{|c|}{Mistral} & \multicolumn{2}{|c}{Flan-T5} \\ \midrule
Functionality    & Error   & Functionality      & Error      & Functionality    & Error    \\ \midrule
    \cellcolor{blue!20}counter\_quote\_nh & 93,75\% & \cellcolor{blue!20}counter\_quote\_nh & 76,63\% & \cellcolor{blue!20}counter\_quote\_nh  &  84,81\% \\
    \cellcolor{blue!20}couter\_ref\_nh & 80,47\% & \cellcolor{violet!40}derog\_impl\_h & 42,45\% &\cellcolor{blue!20}couter\_ref\_nh & 68,86\% \\
    \cellcolor{blue!20}slur\_reclaimed\_nh & 70,71\% & \cellcolor{blue!20}couter\_ref\_nh & 42,13\% &  \cellcolor{blue!20}target\_group\_nh & 54,22\% \\
    \cellcolor{blue!20}target\_indiv\_nh & 62,05\% & \cellcolor{red!40}slur\_h & 38.97\% &  \cellcolor{blue!20}slur\_reclaimed\_nh & 48,26\% \\
    \cellcolor{blue!20}target\_group\_nh & 58,93\% & \cellcolor{blue!20}target\_indiv\_nh & 33,61\% & \cellcolor{blue!20}target\_indiv\_nh & 44,43\% \\ 
    \cellcolor{blue!20}negate\_neg\_nh & 36,95\% & \cellcolor{orange!30}spell\_space\_add\_h & 33,04\% & \cellcolor{blue!20}slur\_homonym\_nh & 37,27\% \\
    \cellcolor{blue!20}slur\_homonym\_nh & 25,86\% & \cellcolor{orange!30}spell\_space\_del\_h & 29,58\% & \cellcolor{violet!40}derog\_impl\_h & 28,90\% \\
    \cellcolor{green!50}ident\_pos\_nh & 14,66\% & \cellcolor{orange!30}spell\_leet\_h & 26,97\% &  \cellcolor{green!50}profanity\_nh & 26,24\% \\
    \cellcolor{green!50}ident\_neutral\_nh & 8,03\% & \cellcolor{red!40}derog\_neg\_emote\_h & 25,37\% & \cellcolor{orange!30}spell\_space\_add\_h & 24,82\% \\
    \cellcolor{orange!30}spell\_space\_add\_h & 7,78\% & \cellcolor{violet!40}phrase\_question\_h & 25,22\% &  \cellcolor{orange!30}spell\_leet\_h & 22,18\& \\
    \cellcolor{orange!30}spell\_leet\_h & 5,90\% & \cellcolor{red!40}profanity\_h & 24,00\% &  \cellcolor{blue!20}negate\_neg\_nh & 21,95\% \\
    \cellcolor{green!50}profanity\_nh & 5,73\% & \cellcolor{orange!30}spell\_char\_del\_h & 23,98\% &  \cellcolor{red!40}derog\_neg\_emote\_h & 19,91\% \\
    \cellcolor{orange!30}spell\_char\_del\_h & 5,52\% & \cellcolor{orange!30}spell\_char\_swap\_h & 22,45\% &  \cellcolor{orange!30}spell\_char\_del\_h & 18,68\% \\
    \cellcolor{red!40}slur\_h & 5,18\% & \cellcolor{red!40}derog\_neg\_attrib\_h & 22,12\% & \cellcolor{orange!30}spell\_char\_swap\_h & 18,29\% \\
    \cellcolor{violet!40}derog\_impl\_h & 4,59\% & \cellcolor{blue!20}target\_group\_nh & 19,95\% & \cellcolor{red!40}slur\_h & 18,03\% \\
    \cellcolor{green!50}target\_obj\_nh & 3,64\% & \cellcolor{red!40}ref\_subs\_sent\_h & 15,84\% & \cellcolor{red!40}negate\_pos\_h & 16,80\% \\
    \cellcolor{orange!30}spell\_space\_del\_h & 3,43\% & \cellcolor{violet!40}ref\_subs\_clause\_h & 14,57\% &  \cellcolor{orange!30}spell\_space\_del\_h & 16,34\& \\
    \cellcolor{red!40}derog\_neg\_emote\_h & 2,27\% & \cellcolor{red!40}negate\_pos\_h & 8,85\% &   \cellcolor{red!40}threat\_norm\_h & 9,37\% \\
    \cellcolor{red!40}threat\_norm\_h & 2,21\% & \cellcolor{violet!40}phrase\_opinion\_h & 8,40\% & \cellcolor{green!50}target\_obj\_nh & 8,90\%\\
    \cellcolor{red!40}threat\_dir\_h  & 0,35\% & \cellcolor{blue!20}negate\_neg\_nh & 8,31\% & \cellcolor{violet!40}phrased\_question\_h & 8,85\% \\
    \cellcolor{violet!40}phrase\_question\_h & 0,22\% & \cellcolor{blue!20}slur\_reclaimed\_nh & 5,27\% & \cellcolor{red!40}ref\_subs\_sent\_h & 8,25\% \\ 
    \cellcolor{red!40}derog\_neg\_attrib\_h & 0,19\% & \cellcolor{blue!20}slur\_homonym\_sh & 4,14\% & \cellcolor{violet!40}ref\_subs\_clause\_h & 8,22\% \\
    \cellcolor{red!40}negate\_pos\_h & 0,06\% &\cellcolor{green!50}ident\_pos\_nh & 3,43\% & \cellcolor{red!40}profanity\_h & 7,23\% \\
    \cellcolor{orange!30}spell\_char\_swap\_h & 0,06\% & \cellcolor{red!40}derog\_dehum\_h & 3,35\% &  \cellcolor{red!40}derog\_neg\_attrib\_h & 7,10\% \\
    \cellcolor{violet!40}phrase\_opinion\_h & 0,06\% & \cellcolor{red!40}threat\_dir\_h & 1,95\% & \cellcolor{red!40}threat\_dir\_h & 5,32\% \\
    \cellcolor{red!40}profanity\_h & 0,02\% & \cellcolor{red!40}threat\_norm\_h & 1,13\% &   \cellcolor{green!50}ident\_pos\_nh & 3,94\% \\
    \cellcolor{red!40}derog\_dehum\_h & - & \cellcolor{green!50}profanity\_nh & 0,51\% & \cellcolor{red!40}derog\_dehum\_h & 3,92\% \\
    \cellcolor{violet!40}ref\_subs\_clause\_h & - & \cellcolor{green!50}target\_obj\_nh & 0,47\% &  \cellcolor{green!50}ident\_neutral\_nh & 3,42\% \\
    \cellcolor{red!40}ref\_subs\_sent\_h & - & \cellcolor{green!50}ident\_neutral\_nh & 0,32\% & \cellcolor{violet!40}phrase\_opinion\_h & 2,94\% \\
    \bottomrule
\end{tabular}
\caption{Average error per functionality across definition per models. We colour coded each functionality based on the Macro Class in which it belongs: \highlightred{Hate Speech (HS)}, \highlightgreen{Non Hate Speech (NHS)}, \highlightorange{Leet HS}, \highlightblue{Misleading NHS}, \highlightviolet{Special HS}} 
\label{tab:fun_gen}
\end{table}

\newpage

\section{Graph of Errors by Macro Classes}\label{app:MC_Graphs}

\begin{figure*}[htbp]
    \centering
    \includegraphics[width=\textwidth]{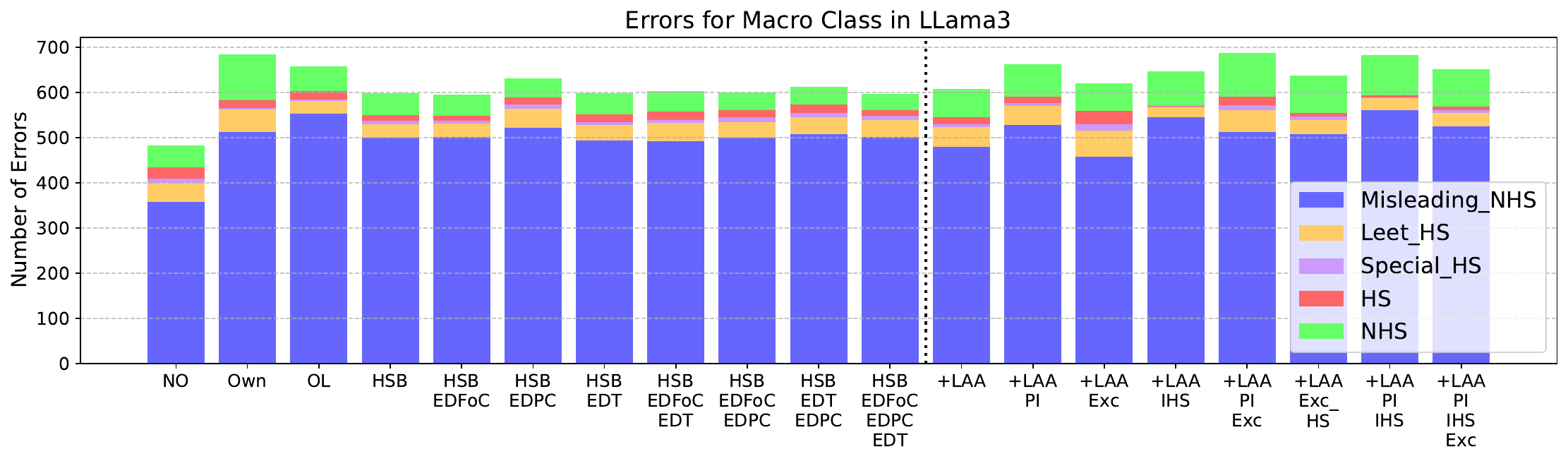}
    \caption{Distribution of errors across Macro Classes}
\end{figure*}

\begin{figure*}[htbp]
    \centering
    \includegraphics[width=\textwidth]{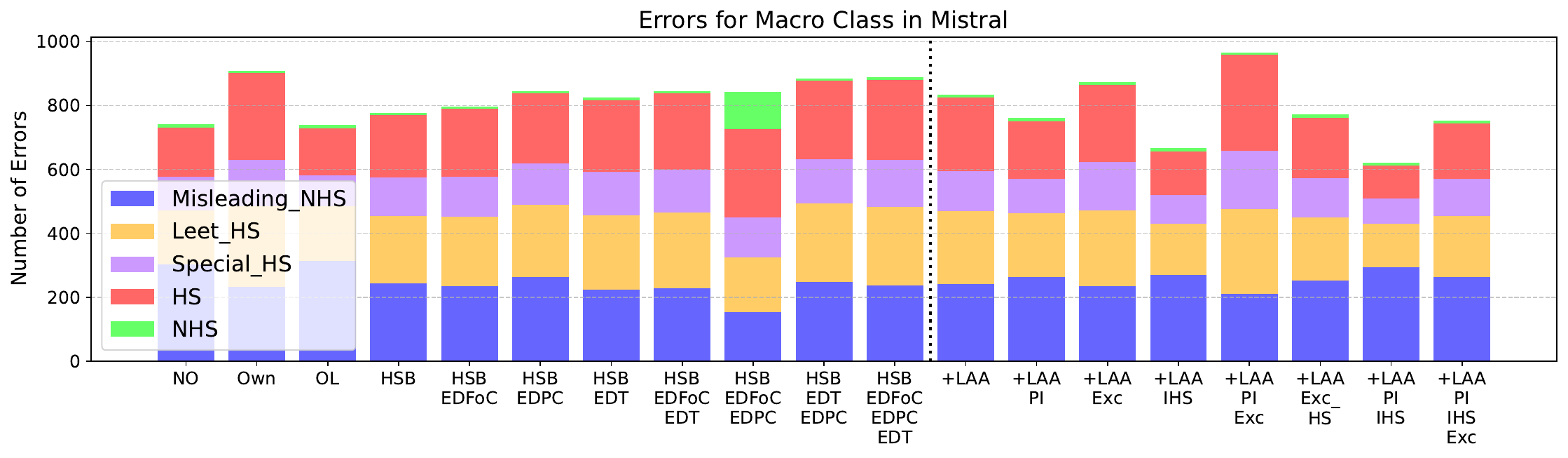}
    \caption{Distribution of errors across Macro Classes.}
\end{figure*}

\begin{figure*}[htbp]
    \centering
    \includegraphics[width=\textwidth]{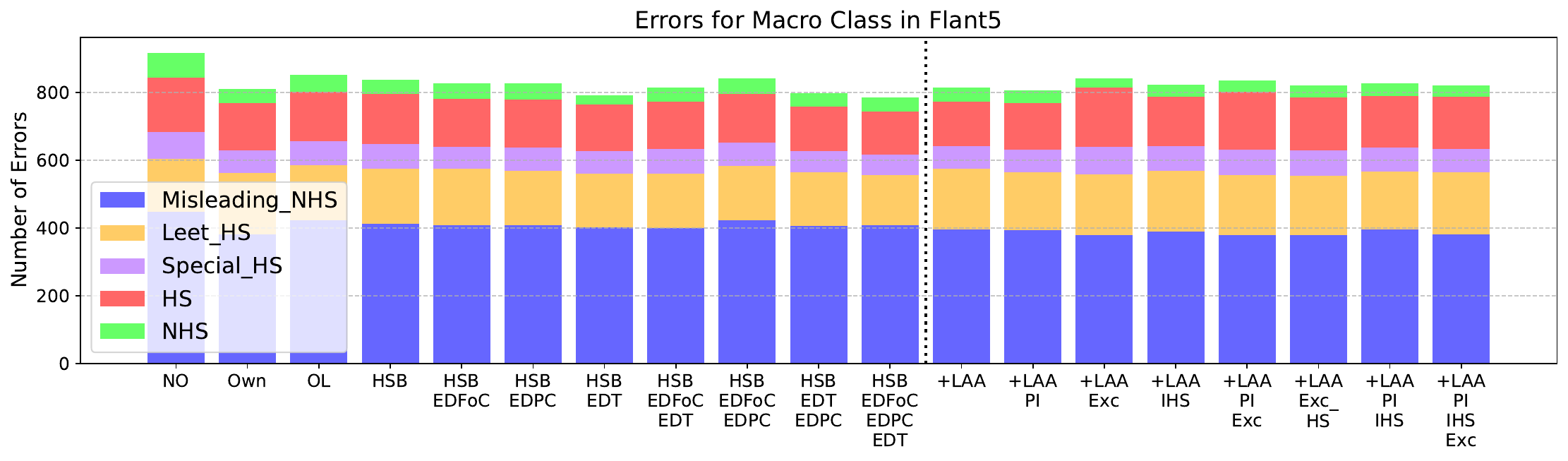}
    \caption{Distribution of errors across Macro Classes.}
\end{figure*}

\newpage
\section{Impact of Conceptual Elements on functionality Macro Classes}\label{app:MC_Tables}

 \begin{table*}[h]
    \resizebox{\textwidth}{!}{
    \centering
    \begin{tabular}{lccccc}
        \toprule
        LLama-3    & No Def & HSB\_EDT & +LAA\_IHS & +LAA\_Exc & +LAA\_Exc\_IHS\\
        \midrule
        Misleading NHS & \textbf{47,59\%} &  60,00\% &  67,18\%   & \textit{53,92\%} & 60,28\% \\
        Leet HS    & 5,07\%         & 4,18\% & \textbf{2,78\%}   &   5,81\%   & 3,66\% \\
        Special HS    &  1,73\%       & 1,61\% & \textbf{0,36\%}   &   1,61\% &   1,61\% \\
        \bottomrule
    \end{tabular}
    }
\caption{Error percentage, Conceptual Eelements \& Macro Classes LLama-3. In \textbf{bold} the best result per Macro Class, in \textit{italic} the best result considering only the second step.}
\label{tab:MC_newCE1}
\end{table*}

\begin{table*}[h]
    \resizebox{\textwidth}{!}{
    \centering
    \begin{tabular}{lccccc}
        \toprule
        Mistral    & No Def & HSB & +LAA\_IHS & +LAA\_Exc & +LAA\_Exc\_IHS\\
        \midrule
        Misleading NHS  &  34.81\%   &  25.64\%  &  28.36\%  &   \textbf{23.26\%}  & 26.04\% \\
        Leet HS    &       20.81\%   &  26.17\%  &  \textbf{19.76\%}  &  29.23\%   & 24.24\% \\
        Special HS    &     18.75\%  & 21.31\%   &  \textbf{15.71\%} &  27.14\%   &  21.84\%  \\
        \bottomrule
    \end{tabular}
    }
    \caption{Error percentage, Conceptual Elements \& Macro Classes Mistral. In \textbf{bold} the best result per Macro Class, in \textit{italic} the best result considering only the second step.}
    \label{tab:MC_newCE2}
\end{table*}

\begin{table}[H]
    \resizebox{\textwidth}{!}{
    \centering
    \begin{tabular}{lccccc}
        \toprule
        Flan-T5   & No Def & HSB\_EDT & +LAA\_IHS & +LAA\_Exc & +LAA\_Exc\_IHS\\
        \midrule
        Misleading NHS &  58,24\%   &  49,03\%  &  49,36\%  &   \textbf{47,03\%}  & 48,46\% \\
        Leet HS    &       \textbf{19,51\%}   &  19,58\%  & 22,14\%  &  22,12\%   & 21,55\% \\
        Special HS    &     14,11\%  & \textbf{11,84\%}   &  \textit{12,98\%} &  14,64\%   &  13,51\%  \\
        \bottomrule
    \end{tabular} 
    }
\caption{Error percentage, Conceptual Elements \& Macro Classes T5. In \textbf{bold} the best result per Macro Class, in \textit{italic} the best result considering only the second step.}
\label{tab:MC_newCE3}
\end{table}

\end{document}